\documentclass[a4paper,fleqn]{cas-dc}

\usepackage[numbers]{natbib}

\usepackage{amsmath,amssymb,amsfonts}
\usepackage{amsthm}
\usepackage{algorithmic}
\usepackage{graphicx}
\usepackage{textcomp}
\usepackage{caption}
\usepackage{subcaption}
\usepackage{etoolbox}
\usepackage{multicol}
\usepackage{url}
\usepackage{tabularray}
\usepackage[linesnumbered,ruled,vlined]{algorithm2e}
\usepackage{comment}
\usepackage{hyperref} 
\usepackage{nomencl}
\usepackage{framed}
\usepackage{booktabs}
\usepackage{array}
\usepackage{bm}

\usepackage{multirow}
\usepackage{threeparttable}

\providecommand{\skinf}{\phantom{0:}}
\providecommand{\skzer}{\phantom{0}}

\ExplSyntaxOn
\bool_gset_true:N \g_stm_nologo_bool
\cs_set:Npn \__first_footerline:
  { \group_begin: \small \sffamily \__short_authors: \group_end: }
\ExplSyntaxOff
\hypersetup{pdftitle={Solver-Free Decision-Focused Learning via Neural Optimization Layers for Edge Selection}}

\begin{document}
\let\WriteBookmarks\relax
\def\floatpagepagefraction{1}
\def\textpagefraction{.001}
\shorttitle{}
\shortauthors{X. Zheng, F. de Nijs, and E. Lam}

\title[mode = title]{Solver-Free Decision-Focused Learning via Neural Optimization Layers for Edge Selection}

\author[1,2]{Xinda Zheng}[orcid=0009-0005-8799-8946]
\cormark[1]
\ead{Xinda.Zheng@monash.edu}

\author[1,2]{Frits de Nijs}[orcid=0000-0003-4466-2447]

\author[1,2]{Edward Lam}[orcid=0000-0002-4485-5014]

\affiliation[1]{organization={Department of Data Science and AI, Faculty of Information Technology, Monash University},
    city={Clayton},
    postcode={3800}, 
    state={Victoria},
    country={Australia}}

\affiliation[2]{organization={ARC Training Centre in Optimisation Technologies, Integrated Methodologies, and Applications (OPTIMA)},
    city={Melbourne},
    postcode={3000}, 
    state={Victoria},
    country={Australia}}

\cortext[cor1]{Corresponding author}

\begin{abstract}
Decision-focused learning trains prediction models to improve downstream decisions, but many methods require costly solver calls or optimal-solution labels. We propose a solver-free decision-focused learning framework for NP-hard edge-selection problems with uncertain parameters in the objective, the constraints, or both. A neural optimization layer is first trained without solution labels, using problem objectives and constraint penalties. Graph-informed edge scoring and iterative refinement encourage concentrated and structurally consistent continuous decisions. The layer is then frozen, and a prediction model is trained through its continuous output using a problem-specific loss evaluated under observed parameters. At deployment, the predictor, optimization layer, and a lightweight decoder produce decisions without solving the original NP-hard problem. Experiments on the asymmetric traveling salesperson and endpoint-free orienteering problems show consistently better decision quality than prediction-focused learning and improvements over most evaluated decision-focused baselines. Our framework requires less training time than the evaluated decision-focused baselines, even when optimization-layer pretraining is included. Neural deployment also produces decisions faster than solver-based deployment, offering a better quality-efficiency trade-off. The layer also reduces training cost in existing decision-focused methods when used as an approximate optimization oracle.
\end{abstract}


\maketitle
\hypersetup{pdfauthor={Xinda Zheng, Frits de Nijs, Edward Lam},pdfsubject={Decision-focused learning}}

\section{Introduction}
\label{sec:introduction}

Many real-world optimization tasks involve making decisions before all relevant information is available. In the predict-then-optimize paradigm, unknown problem parameters are estimated from available context, and an optimization problem is then solved using the estimates \citep{sadana2025survey}. For example, a route may be planned a day in advance using predicted travel times. These uncertain parameters may affect the objective, the constraints, or both: travel times can determine a route's total cost as well as whether it satisfies a travel budget. However, predictors are typically trained to minimize statistical losses such as mean squared error, which measure prediction errors without accounting for their impact on downstream decisions. A large error may leave the optimal decision unchanged, while a small error near a decision boundary can lead to a decision that performs poorly under the true parameters \citep{mandi2024decision}. Decision-focused learning (DFL) addresses this mismatch by training predictors to improve downstream decision quality \citep{wilder2019melding}.

A common approach in DFL achieves this alignment by defining the training loss through solutions to the downstream optimization problem. Evaluating these losses and their gradients requires repeated solver calls \citep{mandi2020smart,vlastelica2020differentiation}, which can be particularly costly for NP-hard edge-selection problems such as the traveling salesperson and orienteering problems. Other approaches avoid repeatedly solving the original problem during predictor updates but require optimal reference solutions \citep{tang2024cave,berden2025solverfree}. However, real-world datasets often contain observed parameters rather than corresponding optimal solutions, so these reference solutions must be computed through potentially costly optimization. At deployment, solver-based methods must still solve an optimization problem for each new prediction \citep{kotary2024learning}, which can hinder applications requiring frequent or time-sensitive decisions.

Reducing computational cost is only part of the challenge. Decision feedback must also account for uncertainty in the constraints. DFL formulations that assume a fixed feasible set do not directly address prediction errors that change decision feasibility \citep{elmachtoub2022smart,vlastelica2020differentiation,tang2024cave}. In edge-selection problems with uncertain constraint parameters, an edge selection that is feasible and yields a favorable objective value under predicted parameters may violate the actual constraints. Some existing approaches incorporate the consequences of infeasibility into their training losses \citep{hu2023twostage,mandi2025odece}, but still rely on optimization solves to obtain the decisions used in training.

A differentiable neural approximation of the parameter-to-decision mapping can support efficient decision feedback and fast decision generation without repeated optimization \citep{cristian2023endtoend}. When the approximation is differentiable with respect to predicted constraint parameters, losses evaluated under realized parameters can provide feedback on both objective performance and constraint violations \citep{cristian2025efficient}. Self-supervised neural optimization methods show that continuous decision representations can be learned from problem objectives and constraints without optimal-solution labels \citep{liu2022gon,zheng2026constraint}. In DFL, these representations both guide discrete decision construction and provide predictor-training feedback through losses evaluated before construction. The relationship between continuous objectives and discrete solution quality therefore matters, and depends on the design of the relaxation and rounding procedure \citep{wang2022unsupervised}.

Motivated by these considerations, we propose DFLNOL, a solver-free decision-focused learning framework for NP-hard edge-selection problems with uncertainty in the objective, the constraints, or both. Its self-supervised edge-selection neural optimization layer (NOL) supports predictor training and fast neural deployment without optimal-solution labels or calls to a solver for the original NP-hard problem. We make the following contributions:

\begin{itemize}
\item \textbf{Solver-free decision-focused learning: }
We propose a two-stage pipeline that pretrains the NOL through self-supervision, then trains a predictor through the frozen layer using objective and constraint losses evaluated under observed parameters. At deployment, the predictor, NOL, and a discrete decoder generate decisions without solving the original NP-hard problem.

\item \textbf{Neural optimization layer for decision-focused learning: }
We develop a NOL that combines graph context and current selection features to capture connectivity and competition among edges. Shared iterative refinement, differentiable output transformations, and structural penalties shape a concentrated, structurally consistent continuous representation for predictor training.

\item \textbf{Decision quality and computational efficiency: }
Experiments on asymmetric traveling salesperson (ATSP) and endpoint-free orienteering (EFOP) benchmarks cover objective, constraint, and joint uncertainty. DFLNOL consistently improves decision quality over prediction-focused learning and remains competitive with evaluated solver-based DFL methods, with substantially less training time, including NOL pretraining, and faster neural deployment.

\end{itemize}

\section{Problem formulation}
\label{sec:problem}

We study edge-selection problems whose objective, constraints, or both depend on problem parameters that are unknown when a decision is made. These parameters are predicted from context, and decision quality is evaluated under their realized values.

\subsection{Edge-selection problems}
\label{sec:edge-selection}

Let $G = (V, E)$ be a directed graph with node set $V = \{1, \dots, n\}$ and edge set $E \subseteq V \times V$, and let $m = |E|$. An edge-selection decision is a binary vector $x \in \{0, 1\}^m$, where $x_e = 1$ indicates that edge $e \in E$ is included in the decision.

An instance is specified by $G$ and a parameter vector $u \in \mathbb{R}^d$, formed by concatenating node-level parameters, edge-level parameters, and instance-level parameters, such as rewards, travel costs, and resource limits. With $p_v$ parameters per node, $p_e$ per edge, and $p_g$ at the instance level, the total dimension is $d = np_v + mp_e + p_g$. We leave $G$ implicit in the notation.

Let $c_\gamma(x;u)\leq0$, indexed by a finite set $\Gamma$, encode the constraints on the selected edges, such as degree, connectivity, and resource requirements. The feasible set is
$
    \mathcal{X}(u) = \bigl\{ x \in \{0, 1\}^m \;\big|\;
    c_\gamma(x; u) \le 0 \text{ for all } \gamma \in \Gamma \bigr\},
$
which we assume to be nonempty for every admissible parameter vector $u$. The objective $F(x;u)$ evaluates a selection under these parameters. Given $u$, the deterministic optimization problem is to find an optimal solution
$
    x^\star(u) \in \arg\min_{x \in \mathcal{X}(u)} F(x; u),
$
where a fixed tie-breaking rule selects $x^\star(u)$ when multiple optimal solutions exist.

\subsection{Prediction and optimization}
\label{sec:prediction}

At decision time, the graph and part of the parameter vector are observed. Write $u=(\xi,w)$, where $\xi$ contains the known parameters and $w$ is realized after the decision is committed.

Where $w$ enters the problem determines what is known about feasibility. If it enters only the objective, $\mathcal{X}(u)$ depends on $\xi$ alone and is known at decision time. If it enters the constraints, feasibility also depends on the unknown parameters.

A context vector $z$ contains covariates observed at decision time and used to predict the unknown problem parameters $w$. A prediction model $h_\theta$, with trainable weights $\theta$, maps the context $z$ to an estimate $\hat w_\theta=h_\theta(z)$ of the unknown problem parameters $w$.

The available historical dataset is
$
    \mathcal{S} = \bigl\{ (u^{(i)}, z^{(i)}) \bigr\}_{i=1}^{N},
$
where each sample records the context and complete realized parameter vector $u^{(i)}=(\xi^{(i)},w^{(i)})$ of a past instance. The dataset contains no optimal or other reference decisions.

For a new instance with known parameters $\xi$ and context $z$, prediction followed by optimization gives
\begin{equation}
    \hat u_\theta = \bigl(\xi,h_\theta(z)\bigr),
    \qquad
    \hat x_\theta = x^\star(\hat u_\theta).
    \label{eq:deployed}
\end{equation}
Predictions are restricted to admissible parameter values, so the estimated problem has a feasible solution. The planned selection $\hat x_\theta$ therefore belongs to $\mathcal X(\hat u_\theta)$, but need not belong to $\mathcal X(u)$.

Let $\mathcal X_0(\xi)$ denote the binary edge selections satisfying all constraints that are independent of $w$. Since $\mathcal X(\hat u_\theta)\subseteq\mathcal X_0(\xi)$, the planned selection satisfies the known constraints. When $w$ enters only the objective, $\mathcal X_0(\xi)=\mathcal X(u)$ and this also ensures feasibility under the realized parameters. Otherwise, the planned selection may require correction.

\subsection{Decision-focused learning problem}
\label{sec:dfl}

When uncertain parameters enter the constraints, we consider problems equipped with a prescribed correction rule $\operatorname{corr}(\,\cdot\,;u):\mathcal X_0(\xi)\longrightarrow\mathcal X(u)$. This rule maps a planned edge selection to an executable selection under the realized parameters, respecting the application's restrictions on changes to the plan. It leaves an already feasible selection unchanged: $\operatorname{corr}(x;u)=x$ for every $x\in\mathcal X(u)$. The rule is part of the problem specification, and its form depends on the application. When uncertainty affects only the objective, no correction is needed and the rule is the identity map.

For a planned selection $x\in\mathcal X_0(\xi)$, we define \emph{regret} by comparing the objective value after correction with the optimal value under the realized parameters:
\begin{equation}
    \begin{aligned}
        \operatorname{Regret}(x;u)
        ={}&F\bigl(\operatorname{corr}(x;u);u\bigr)%
        -F\bigl(x^\star(u);u\bigr).
    \end{aligned}
    \label{eq:regret}
\end{equation}
Regret is nonnegative because the corrected selection is feasible, and it is zero exactly when that selection is optimal. For an already feasible $x$, it reduces to $F(x;u)-F(x^\star(u);u)$. Thus, the consequences of correction are measured through the objective value of the resulting selection.

Decision-focused learning seeks predictions that induce low regret under the joint distribution of $(u,z)$. Its empirical objective on $\mathcal S$ is $\min_\theta\frac1N\sum_{i=1}^{N}\operatorname{Regret}\bigl(\hat x_\theta^{(i)};u^{(i)}\bigr)$, where $\hat x_\theta^{(i)}$ is the planned selection defined by~\eqref{eq:deployed} for sample $i$. The learning task is to use contexts and realized parameters, without reference decisions, to obtain predictions whose induced edge selections perform well after any required correction.

\section{Related Work}
\label{sec:related-work}

We organize related work around three connected questions: how predictors receive decision feedback, what a learned optimization mapping contributes, and how neural representations of discrete solutions can support predictor training.

\subsection{Training feedback in decision-focused learning}
\label{subsec:dfl}

DFL methods obtain training feedback through different treatments of the downstream optimization problem. Differentiable optimization layers compute sensitivities through continuous problems \citep{amos2017optnet,agrawal2019differentiable}. For combinatorial problems, regularized relaxations provide differentiable approximations \citep{wilder2019melding}, while MIPaaL tightens a continuous relaxation using cutting planes \citep{ferber2020mipaal}. DYS-Net improves the efficiency of differentiable optimization by combining Davis--Yin splitting for quadratically regularized relaxations with Jacobian-free backpropagation \citep{mckenzie2024differentiating}. Building on this approach, \citet{mandi2025surrogate} show that quadratic smoothing can leave large regions with zero decision-loss gradients and combine differentiable optimization with surrogate losses to improve predictor training. Oracle-based methods instead construct training signals from returned solutions, as in black-box differentiation and differentiable perturbations \citep{vlastelica2020differentiation,berthet2020learning}. SPO+ and perturbation-gradient losses similarly use optimization oracles to evaluate surrogate losses and their gradients \citep{elmachtoub2022smart,gupta2024directional}. These approaches obtain feedback by computing solutions to original or relaxed optimization problems during training. Repeated optimization can remain costly, particularly when the underlying oracle solves an NP-hard problem \citep{mandi2020smart}.

Several approaches reduce the cost of obtaining decision feedback. Relaxations and warm starts accelerate optimization calls \citep{mandi2020smart}, while contrastive and ranking losses reuse candidate solutions through caching \citep{mulamba2021contrastive,mandi2022decision}. Geometric losses further reduce online optimization: CaVE uses quadratic-program projections or their approximations to align predicted costs with a cone associated with an optimal reference solution \citep{tang2024cave}, and LAVA computes a solver-free training loss from an optimal vertex and its precomputed adjacent vertices \citep{berden2025solverfree}. Beyond reusing solutions or exploiting their geometry, learned loss surrogates approximate decision losses using targets obtained from optimization evaluations. These include locally optimized decision losses and their feature-based extensions \citep{shah2022decision,shah2024leaving}, as well as LANCER, which alternates between surrogate fitting and predictor updates \citep{zharmagambetov2023landscape}. These approaches reduce the computation required for predictor feedback, but do not themselves learn a solution mapping for deployment.

The content of the feedback also changes when unknown parameters enter constraints. \citet{hu2023packing} incorporate solution correction and modification costs into training for packing and covering linear programs. Their subsequent MILP framework formulates correction as a second-stage optimization problem \citep{hu2023twostage}. Odece instead penalizes predictions that induce infeasible solutions or exclude true optimal solutions, using optimization-generated candidates and optimal reference solutions \citep{mandi2025odece}. These studies establish different ways to account for realized feasibility. Our framework evaluates objective and constraint losses on continuous selections produced by a learned mapping, with discrete construction and execution correction outside backpropagation. The mapping therefore provides an interface for both training feedback and deployed solution generation.

\subsection{Learned optimization mappings for decision-focused learning}
\label{subsec:learned-modules}

Optimization proxies learn solution mappings from supplied problem parameters. DC3 and primal-dual learning demonstrate self-supervised approximation for continuous constrained optimization without solution labels \citep{donti2021dc3,park2023selfsupervised}. ProjectNet extends this approach to predictor training by first learning an optimization module and then training a predictor through the pretrained mapping \citep{cristian2023endtoend}. Its learned updates use approximate projections onto convex feasible regions, with applications including shortest path and matching. The extended framework also considers two-stage stochastic optimization and differentiation with respect to constraint parameters \citep{cristian2025efficient}. An alternative is LtOF, which directly maps context to solutions and is evaluated on continuous optimization problems with objective uncertainty and fixed feasible sets \citep{kotary2024learning}. We retain explicit parameter predictions, allowing the same predictor to support neural or solver-based deployment. Within this interface, we design the NOL to represent interacting edge choices and support discrete solution construction for NP-hard edge-selection problems with objective and constraint uncertainty.

Neural combinatorial optimization offers several representations, with different requirements for integration into predictor training. DIFUSCO learns a diffusion model from optimal or heuristic solution vectors \citep{sun2023difusco}, requiring a solution corpus unavailable in our setting. Attention-based routing models avoid solution labels by using policy gradients over discrete construction trajectories \citep{kool2019attention}. DIMES introduces a continuous parameterization of solution distributions but retains REINFORCE-based sampling for training and fine-tuning \citep{qiu2022dimes}. Thus, continuous parameterization does not necessarily imply direct differentiation of a loss on the represented selection. Our NOL uses this latter mechanism, allowing predictor gradients to pass through its continuous output without differentiating discrete construction.

Self-supervised relaxation methods provide mechanisms for learning continuous representations directly from problem structure. \citet{wang2022unsupervised} establish conditions under which entry-wise concave relaxations and sequential rounding yield feasible solutions with objective bounds. GON learns node assignments through constraint-aware losses for balanced graph partitioning and maximum independent set \citep{liu2022gon}. These methods address solution learning and recovery for supplied instances. Their use in contextual edge selection additionally requires a representation of edge connectivity and competition, together with losses that train parameter predictions according to realized objective performance and feasibility.

The edge-selection model of \citet{zheng2026constraint} provides our direct architectural foundation. It combines a cost-attention encoder, self-supervised objective and constraint losses, and lightweight decoding to generate solutions from supplied parameters. We build on this foundation with decision-dependent edge scoring, shared iterative refinement, and problem-specific output transformations. The scorer incorporates relationships derived from the current selection, so each refinement round evaluates edges in the context of surrounding choices. These relationships become part of the continuous computation through which the predictor learns. Our NOL thus extends the earlier solution-generation model into a structured optimization layer for DFL, while retaining discrete decoding for fast deployment.

\section{Solver-free decision-focused learning framework}
\label{sec:framework}

This section presents the framework, explains its two training stages, and describes how the trained models produce decisions at deployment.

\subsection{Framework overview}
\label{sec:overview}

The decision rule in~\eqref{eq:deployed} converts predictions into decisions through the solution map $u\mapsto x^\star(u)$. Evaluating this map requires solving the underlying combinatorial optimization problem, as in solver-based DFL methods such as SPO~\citep{elmachtoub2022smart}. Its discrete output also obstructs direct differentiation: changes in the predicted parameters can leave the selected decision unchanged, giving zero gradients wherever the map is locally constant. We replace this map with a learned NOL that produces a continuous edge-selection representation.

Given context $z$, the predictor $h_\theta$ estimates the unknown parameters as $\hat w_\theta=h_\theta(z)$, forming $\hat u_\theta=(\xi,\hat w_\theta)$ together with the known parameters. The NOL, with weights $\omega$, returns $P=\tilde\pi_\omega(\hat u_\theta)\in[0,1]^m$, where each entry indicates the degree to which an edge is selected. Differentiability with respect to the NOL input allows its output to guide predictor training. A problem-specific decoder converts the continuous representation into a discrete decision at deployment. The NOL architecture is described in Section~\ref{sec:optimization-layer}.

\begin{figure*}[H]
  \centering
  \includegraphics[width=\textwidth]{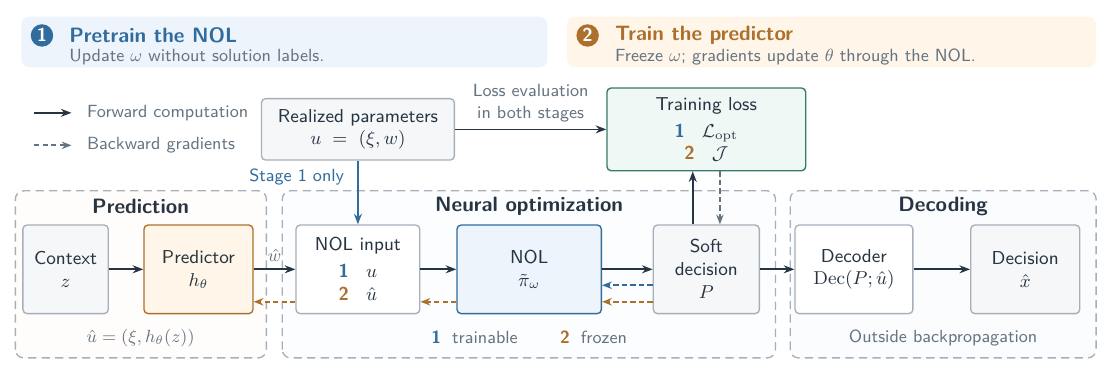}
  \caption{Two-stage training and deployment with a shared neural optimization layer (NOL). Numbered inputs and losses correspond to the two training stages. Intermediate NOL quantities used by the losses are omitted. Deployment follows the horizontal path.}
  \label{fig:framework-overview}
\end{figure*}

Learning proceeds in two stages (Figure~\ref{fig:framework-overview}). Stage 1 trains the NOL on realized parameter vectors using the problem objective and structural requirements. Stage 2 freezes the NOL and trains the predictor through it: predicted parameters generate the continuous selection, while the realized parameters recorded in the training data evaluate its objective performance and any selected constraint violations or structural adjustments. Gradients update the predictor without decoding the selection.

Both stages use the historical data $\mathcal S$ and the problem specification, without reference decisions or optimal objective values. Model selection compares decoded decision quality on validation data. Deployment combines the trained networks with decoding and, where needed, a solver-free execution rule. The framework therefore avoids calls to the original combinatorial optimization solver during learning and deployment; solver-based methods and reference solutions are used only for benchmarking.

\subsection{Training the neural optimization layer}
\label{sec:module}

For each realized parameter vector $u$, the NOL produces $P_\omega=\tilde\pi_\omega(u)$. Following the constraint-guided self-supervised approach of \citet{zheng2026constraint}, Stage 1 seeks both good objective values and selections consistent with the problem structure. We formulate this as constrained learning, with penalty strengths adjusted according to the deviations observed during training.

Let $\tilde F(P;u)$ be a differentiable extension of the objective that agrees with $F$ on binary decisions. For a linear objective, $\tilde F(P;u)=\sum_{e\in E}c_e(u)P_e$, with coefficients obtained from the supplied parameters. The objective alone need not favor meaningful selections: selecting no edges, for example, gives zero cost when edge costs are positive, even if the problem requires a nonempty selection.

Training losses may use both the edge-selection output and differentiable intermediate quantities computed in the same NOL evaluation. We leave these intermediate quantities implicit and write the losses as functions of the network parameters and supplied data, rather than of $P$ alone. This introduces no additional data requirement.

Let $\mathcal K$ index structural requirements, and let $v_k(\omega;u)\geq0$ measure the deviation from requirement $k$ in the NOL computation on $u$, with zero indicating that the condition is satisfied. These differentiable or piecewise-differentiable measures can describe relaxed degree or resource constraints, or proximity to binary selections. They are normalized by problem-dependent scales; for example, budget excess can be expressed relative to the available budget, and local residuals can be averaged over nodes.

For each requirement, a tolerance $\varepsilon_k\geq0$ specifies the acceptable average deviation, allowing for numerical and representational limitations. Additional structural preferences without a justified tolerance are included in a differentiable regularizer $\mathcal R(\omega;u)$ with fixed coefficients. Unlike the deviation measures, this regularizer need not be nonnegative or vanish on feasible binary decisions.

Positive data-dependent normalization factors may also be applied to the training objective; they are fixed with respect to the trainable parameters and left implicit below. The loss definitions, scales, and any instance-dependent weighting are specified in the case studies.

We reserve a validation subset of $\mathcal S$ for model selection. With $\mathbb E_{\mathrm{tr}}$ denoting the empirical average over the remaining training instances, the NOL training problem is
\begin{equation}
    \begin{aligned}
        \min_\omega\quad&
        \mathbb{E}_{\mathrm{tr}}
        \bigl[
            \tilde{F}(P_\omega;u)
            +\mathcal{R}(\omega;u)
        \bigr] \\
        \text{subject to}\quad&
        \mathbb{E}_{\mathrm{tr}}
        \bigl[v_k(\omega;u)\bigr]
        \leq\varepsilon_k,
        \quad k\in\mathcal{K}.
    \end{aligned}
    \label{eq:stage-one}
\end{equation}
These conditions constrain average training deviations, not the feasibility of every individual output.

Associating a nonnegative multiplier $\lambda_k$ with each condition gives the per-instance Lagrangian loss
\begin{equation}
    \begin{aligned}
        \mathcal{L}_{\mathrm{opt}}(\omega;u,\lambda)
        ={}& \tilde{F}(P_\omega;u)
             +\mathcal{R}(\omega;u) \\
        &+\sum_{k\in\mathcal{K}}
            \lambda_k
            \bigl(v_k(\omega;u)-\varepsilon_k\bigr).
    \end{aligned}
    \label{eq:loss}
\end{equation}
For fixed multipliers, the tolerance terms do not affect the network gradient, but they determine the multiplier updates.

Training uses stochastic primal-dual updates. For a mini-batch $\mathcal B_t$, let $\widehat{\mathcal L}_t(\omega,\lambda)$ be the batch average of~\eqref{eq:loss} and define $\bar v_{k,t}=\frac{1}{|\mathcal B_t|}\sum_{u\in\mathcal B_t}v_k(\omega_t;u)$. The network follows descent and the multipliers follow projected ascent:
\begin{equation}
    \begin{aligned}
        \omega_{t+1}
        &=
        \omega_t
        -
        \eta_\omega
        \nabla_\omega
        \widehat{\mathcal{L}}_t(\omega_t,\lambda_t), \\
        \lambda_{k,t+1}
        &=
        \Bigl[
            \lambda_{k,t}
            +\eta_{\lambda,k}
            \bigl(\bar v_{k,t}-\varepsilon_k\bigr)
        \Bigr]_+,
        \quad k\in\mathcal{K},
    \end{aligned}
    \label{eq:layer-primal-dual}
\end{equation}
where $[a]_+=\max\{a,0\}$, $\eta_\omega>0$ is the network step size, and $\eta_{\lambda,k}>0$ allows a separate step size for each multiplier. An adaptive gradient optimizer may replace the network descent step. Multipliers are held fixed for that step, and their update history is not differentiated through.

Each multiplier increases when its batch deviation exceeds the tolerance and decreases, subject to nonnegativity, when the deviation falls below it. A zero tolerance therefore permits only increases or no change. Normalizations, tolerances, initial multipliers, and step sizes remain design choices reported in the case studies; these nonconvex updates do not guarantee global optimality or exact feasibility.

After each epoch, outputs on the fixed validation set are decoded under the realized parameters using the procedure in Section~\ref{sec:deployment}. We retain the checkpoint with the lowest mean objective value of the resulting feasible decisions. This avoids comparing training losses whose multiplier values change over time. During this stage, decoding is used for validation, not for gradient updates.

Let $\omega^\star$ denote the selected weights and write $\tilde\pi^\star=\tilde\pi_{\omega^\star}$. These weights are frozen for Stage 2, while differentiation with respect to the NOL inputs remains available.

\subsection{Training the prediction model}
\label{sec:stage-two}

For a training sample $(u,z)$ with $u=(\xi,w)$, the predictor supplies $\hat u_\theta=(\xi,h_\theta(z))$ to the frozen NOL, producing $P_\theta=\tilde\pi^\star(\hat u_\theta)$. Predictions generate the continuous selection; the realized parameters already recorded in the training data determine how it is evaluated.

This distinction matters when uncertainty affects feasibility: a selection satisfying a budget under predicted costs may violate it under realized costs. Stage 1 teaches the NOL to respond to supplied parameters, but cannot eliminate this discrepancy. We therefore define a continuous decision loss $\mathcal J(\hat u;u)$ that evaluates the frozen NOL computation driven by $\hat u$ under the recorded parameters $u$. Predictor training solves $\min_\theta\;\mathbb{E}_{\mathrm{tr}}\bigl[\mathcal{J}(\hat u_\theta;u)\bigr]$. The average is over training pairs $(u,z)$. A general form of the loss, with $P=\tilde\pi^\star(\hat u)$, is
\begin{equation}
    \begin{aligned}
        \mathcal{J}(\hat u;u)
        ={}&\tilde F(P;u)
        +\sum_{k\in\mathcal{K}_{\mathrm{pred}}}
            \mu_k v_k^{\mathrm{pred}}(\hat u;u)\\
        &+\mathcal{R}_{\mathrm{pred}}(\hat u;u).
    \end{aligned}
    \label{eq:prediction-decision-loss}
\end{equation}
Here $v_k^{\mathrm{pred}}(\hat u;u)\geq0$ measures a selected constraint violation, such as budget excess under realized resource consumption, and $\mu_k\geq0$ is its weight. The optional regularizer $\mathcal R_{\mathrm{pred}}$ adjusts the continuous evaluation, for example by reducing reward credited to selections that lack structural support. Both terms may use intermediate quantities from the same NOL computation and are differentiable or piecewise differentiable with respect to its input. The recorded parameters $u$ remain fixed: predictions change what the NOL computes, not the cost or resource coefficients used to evaluate it.

The stages serve different purposes: Stage 1 learns useful selections for supplied parameters, whereas Stage 2 learns predictions that induce good selections under realized parameters. Its penalties are therefore task-specific, not copied automatically from Stage 1, and the Stage 1 multipliers are not transferred. The terms, scales, and coefficients are specified in the case studies. Omitting the additional terms recovers objective-only training, $\mathcal J(\hat u_\theta;u)=\tilde F(P_\theta;u)$.

For one training sample, the chain rule gives
\begin{equation}
    \begin{aligned}
        \nabla_\theta\mathcal{J}(\hat u_\theta;u)
        ={}&
        \left(
            \frac{\partial h_\theta(z)}{\partial\theta}
        \right)^{\!\top}
        \nabla_{\hat w_\theta}\mathcal{J}(\hat u_\theta;u).
    \end{aligned}
    \label{eq:gradient}
\end{equation}
The rightmost derivative is with respect to the predicted component of the first argument of $\mathcal J$. It includes all paths through the frozen NOL, including intermediate quantities used by the loss, rather than only the final $P$. Only $\theta$ is updated; $\omega^\star$, $\xi$, and the evaluation parameters $u$ remain fixed. At nondifferentiable points, a subgradient is used. Neither decoding nor execution correction participates in this computation.

This continuous loss is a surrogate, not an exact reformulation of executed regret or the correction procedure in Section~\ref{sec:dfl}. Its penalties provide a signal about realized violations but do not guarantee feasibility or better decoded decisions. Predictions may also enter regions where the NOL approximation is inaccurate; predictor initialization and output-domain restrictions are specified in the case studies.

Model selection therefore uses executed decision quality. After each epoch, predictions on a fixed validation set pass through the frozen NOL and decoder. We evaluate the resulting decisions under realized parameters, apply the prescribed correction where required, and retain the checkpoint with the lowest mean executed objective value, with parameters denoted by $\theta^\star$. Continuous-output quality and executed quality are assessed separately in the case studies.

\subsection{Deployment}
\label{sec:deployment}

At deployment, a new instance provides $\xi$ and $z$, while $w$ remains unknown. The selected predictor and NOL produce a continuous representation, and a problem-specific decoder $\operatorname{Dec}(P;u)$ constructs a discrete decision under the supplied parameters: $\hat{u}=\bigl(\xi,h_{\theta^\star}(z)\bigr)$, $P=\tilde{\pi}^\star(\hat{u})$, and $\hat{x}=\operatorname{Dec}(P;\hat{u})$. This requires one forward pass through each network and one decoding operation, without parameter updates or solving the original optimization problem.

For the settings considered here, the decoder produces a feasible decision for every continuous input when the supplied parameters satisfy its applicability conditions. Predictions must lie in this admissible domain; the conditions, decoding algorithms, and feasibility arguments are given in the case studies. Under these assumptions, $\hat x\in\mathcal X(\hat u)\subseteq\mathcal X_0(\xi)$. This guarantee comes from the decoder, not from assuming that $P$ is feasible or ranks edges accurately. Edge rankings still affect objective quality, and feasibility does not imply optimality.

When uncertainty affects only the objective, $\hat x$ is also feasible under the realized parameters because the feasible set depends on $\xi$ alone. With uncertain constraints, the rule in Section~\ref{sec:dfl} yields \mbox{$x^{\mathrm{exec}}=\operatorname{corr}(\hat x;u)\in\mathcal X(u)$}, leaving an already feasible decision unchanged. Performance is evaluated by $F(x^{\mathrm{exec}};u)$. Correction is a separate execution step, not part of planning with unknown parameters, and must itself be solver-free for the complete process to avoid solver calls.

Validation in Stage 2 uses this same prediction, decoding, and execution procedure. The case studies specify the execution rules and evaluate how well the continuous training criterion translates into deployed decision quality.

\section{Neural optimization layer for edge selection}
\label{sec:optimization-layer}

This section describes the NOL computation used in Section~\ref{sec:framework}: encoding the input graph, scoring edges using the current selection, and updating that selection through a differentiable transformation.

\subsection{Overall architecture and input representation}
\label{subsec:layer-overview}

For a directed graph $G=(V,E)$, the NOL maps supplied parameters to a continuous edge selection. In the architectural formulas below, $u$ denotes the supplied vector: the realized parameters during Stage 1, replaced by $\hat u_\theta=(\xi,h_\theta(z))$ during predictor training and deployment. The same computation applies in both cases.

The input comprises node features $\mathbf a_i$, edge features $\mathbf a_{ij}$, and, where needed, global features $\mathbf a_G$. Examples include node rewards, edge travel costs, and an instance-wide budget. Node features may also include statistics of incident edges. Any parameter on which the decision should depend must enter these features or the output transformation; supplying it only to the training loss is insufficient.

Preprocessing may use fixed statistics from the NOL training set or positive, instance-dependent scales computed from the supplied input. Instance-dependent normalization remains on the differentiation path, and quantities with common physical units are scaled consistently. These input transformations are distinct from the objective normalization in Section~\ref{sec:module}; their definitions are specified in the case studies.

The output is $P=\tilde\pi_\omega(u)\in[0,1]^m$. To express interactions between edges, we associate it with a matrix $\mathbf P\in[0,1]^{n\times n}$, setting $\mathbf P_{ij}=P_e$ for $e=(i,j)\in E$ and zero otherwise. Conversely, $\operatorname{vec}_E(\mathbf P)$ extracts the edge entries in the ordering of $P$. These values are continuous selection weights, not necessarily marginal probabilities under a distribution over feasible solutions.

Because the continuous selection enters the training criterion, useful edge rankings alone are insufficient: the weights should also represent compatible choices. For a linear objective with coefficient vector $c(u)$, any soft selection $P$ and feasible binary decision $x\in\mathcal X(u)$ satisfy
\begin{equation}
    \left|\tilde F(P;u)-F(x;u)\right|
    \leq \|c(u)\|_\infty\|P-x\|_1.
    \label{eq:layer-objective-discrepancy}
\end{equation}
Proximity to a discrete decision therefore controls the soft-objective error. The bound itself establishes neither feasibility nor optimality; concentration must be considered together with structural consistency and objective quality.

\begin{figure*}[H]
  \centering
  \includegraphics[width=\textwidth]{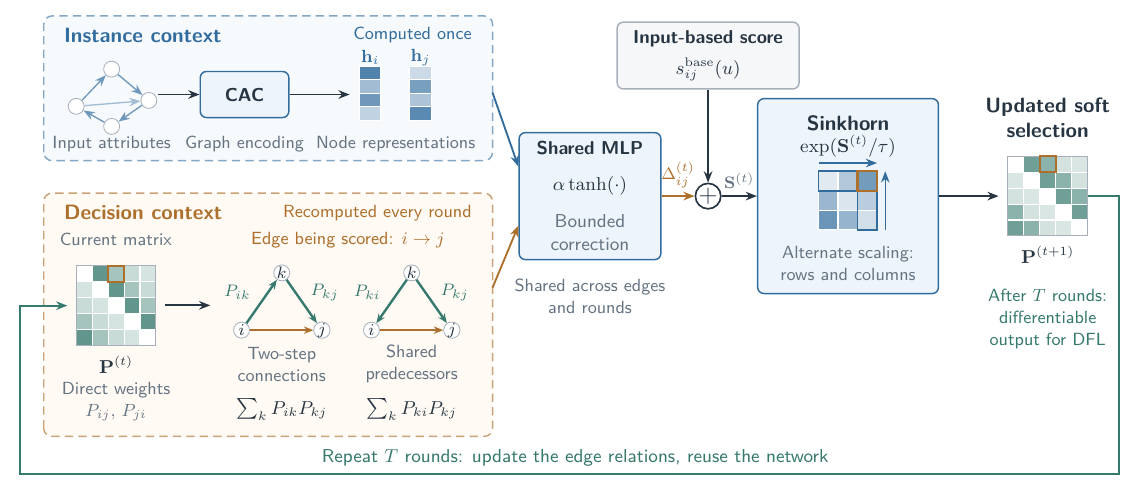}
  \caption{Iterative refinement in the neural optimization layer. Orange highlights the edge $(i,j)$ being scored.}
  \label{fig:nol-mechanism}
\end{figure*}

The architecture combines an instance representation computed once with a selection pattern updated over successive rounds (Figure~\ref{fig:nol-mechanism}). A graph encoder produces node representations, a shared scorer evaluates edges using the current selection, and an output transformation jointly updates their weights.

For a fixed number $T$ of rounds, the computation is
\begin{equation}
    \begin{aligned}
        H &= \operatorname{Enc}_{\omega}(u),\\
        \mathbf{P}^{(0)} &= \mathcal{I}(u),\\
        \mathbf{P}^{(t+1)}
        &= \mathcal{U}_{\omega}\bigl(\mathbf{P}^{(t)},H;u\bigr).
    \end{aligned}
    \label{eq:layer-overall-computation}
\end{equation}
Here $t=0,\ldots,T-1$, $H$ contains the node representations, $\mathcal I$ initializes the soft selection, and $\mathcal U_\omega$ combines scoring and the output transformation. The node representations are computed once and reused, without being detached from gradient computation. The selection and its derived features change in each round, while the update parameters are shared. The returned vector is $\tilde\pi_\omega(u)=\operatorname{vec}_E\bigl(\mathbf P^{(T)}\bigr)$.

\subsection{Cost-attention convolution}
\label{subsec:layer-cac}

Cost-attention convolution (CAC) incorporates the alternatives around each node into its representation by aggregating neighboring features with cost-dependent weights~\citep{zheng2026constraint}.

The initial node representation is $\mathbf h_i^{(0)}=\varphi(W_{\mathrm{in}}\mathbf a_i+\mathbf b_{\mathrm{in}})\in\mathbb R^{d_h}$, where $\varphi(a)=a/(1+\exp(-a))$ is the elementwise SiLU activation.

Let $\bar C_{ij}(u)$ be the normalized cost attribute for edge $(i,j)$. It may represent travel cost or resource consumption and need not equal the objective coefficient $c_e(u)$. For outgoing neighbors $\mathcal N^+(i)=\{j:(i,j)\in E\}$, attention weights are
\begin{equation}
    A_{ij}
    =
    \frac{\exp(-\beta\bar{C}_{ij})}
    {\sum_{k\in\mathcal{N}^{+}(i)}
        \exp(-\beta\bar{C}_{ik})},
    \qquad j\in\mathcal{N}^{+}(i),
    \label{eq:layer-cost-attention}
\end{equation}
with zero weights outside the neighborhood. The learned scalar $\beta$ is initialized positively; when $\beta>0$, lower-cost neighbors receive larger weights. Nodes without outgoing neighbors receive a zero aggregated message.

The message $\mathbf m_i=\sum_{j\in\mathcal N^+(i)}A_{ij}\mathbf h_j^{(0)}$ is combined with the node's own representation as $\mathbf h_i=\varphi(W_{\mathrm{self}}\mathbf h_i^{(0)}+W_{\mathrm{neigh}}\mathbf m_i+\mathbf b_{\mathrm{cac}})$. Stacking these vectors gives $H\in\mathbb R^{n\times d_h}$. The encoder weights, including $\beta$, belong to $\omega$.

Aggregation follows outgoing edges and imposes no symmetry. CAC describes the supplied instance; the scorer below adds information about the network's current selection.

\subsection{Decision-dependent edge scoring}
\label{subsec:layer-scoring}

The scorer combines the graph context with the current soft matrix $\mathbf P^{(t)}$, allowing an edge to be evaluated in light of the choices currently made around it.

For edge $(i,j)$, $\mathbf P_{ij}^{(t)}$ and $\mathbf P_{ji}^{(t)}$ give the direct selection weights in both directions, with zero assigned to an absent reverse edge. Two additional relations describe surrounding choices:
\begin{equation}
    \begin{aligned}
        R_{ij}^{(t)}
        &=\bigl(\mathbf{P}^{(t)}\mathbf{P}^{(t)}\bigr)_{ij}
         =\sum_k\mathbf{P}_{ik}^{(t)}
                 \mathbf{P}_{kj}^{(t)},\\
        M_{ij}^{(t)}
        &=\bigl(\mathbf{P}^{(t)\top}\mathbf{P}^{(t)}\bigr)_{ij}
         =\sum_k\mathbf{P}_{ki}^{(t)}
                 \mathbf{P}_{kj}^{(t)}.
    \end{aligned}
    \label{eq:layer-decision-relations}
\end{equation}
$R_{ij}^{(t)}$ measures two-step connections $i\to k\to j$. $M_{ij}^{(t)}$ measures weight received by $i$ and $j$ from common predecessors, exposing cases where a node spreads its selection across several destinations. These describe the soft selection pattern, not feasibility violations. The scorer learns whether each relation should increase or decrease an edge's score.

The feature vector supplied to the scorer is
\begin{equation}
    \begin{aligned}
        \mathbf{f}_{ij}^{(t)}
        =\operatorname{concat}\bigl(
            &\mathbf{a}_{ij},\mathbf{a}_G,
            \mathbf{h}_i,\mathbf{h}_j,\\
            &\mathbf{P}_{ij}^{(t)},\mathbf{P}_{ji}^{(t)},
            \kappa_{\mathrm{rel}} R_{ij}^{(t)},\kappa_{\mathrm{rel}} M_{ij}^{(t)}
        \bigr),
    \end{aligned}
    \label{eq:layer-edge-features}
\end{equation}
where $\kappa_{\mathrm{rel}}>0$ is a fixed scale for the relation features. Edge and global features describe the supplied parameters, endpoint representations provide graph context, and the remaining entries describe the current selection. The order of $\mathbf h_i$ and $\mathbf h_j$ distinguishes source and destination. Unused feature groups are omitted in the problem-specific instantiation.

A shared MLP computes $q_{ij}^{(t)}=\mathbf w_2^\top\varphi(W_1\mathbf f_{ij}^{(t)}+\mathbf b_1)+b_2$. Its parameters belong to $\omega$ and are shared across edges and rounds. The input dimension depends on the supplied features, but the scoring rule retains the same form.

We combine the learned output with a baseline score derived from the instance:
\begin{equation}
    \begin{aligned}
        \Delta_{ij}^{(t)}
        &=\alpha\tanh\bigl(q_{ij}^{(t)}\bigr),\\
        S_{ij}^{(t)}
        &=s_{ij}^{\mathrm{base}}(u)+\Delta_{ij}^{(t)},
    \end{aligned}
    \label{eq:layer-edge-score}
\end{equation}
where larger $S_{ij}^{(t)}$ indicates a stronger selection preference. A negative normalized cost, for example, provides a baseline for cost minimization; other problems use a baseline suited to their attributes. The bound $|\Delta_{ij}^{(t)}|<\alpha$ controls how far learning can revise this baseline, so $\alpha>0$ is chosen together with the feature scales. Every round corrects the same baseline rather than accumulating corrections onto the previous score.

Learned scores guide the NOL computation; they do not replace the recorded objective or constraint coefficients used to evaluate its outputs in either training stage of Section~\ref{sec:framework}.

\subsection{Differentiable updates and iterative refinement}
\label{subsec:layer-refinement}

Let $\mathbf S^{(t)}$ collect the edge scores. The update $\mathbf P^{(t+1)}=\mathcal B_\tau(\mathbf S^{(t)};u)$ converts them into selection weights in $[0,1]$, with zero entries outside $E$. Here $\mathcal B_\tau$ is a differentiable, problem-specific transformation and $\tau>0$ controls its smoothness.

The case studies use Sinkhorn normalization to implement $\mathcal B_\tau$, with adaptations to each problem (Figure~\ref{fig:nol-mechanism}). Alternating row and column normalization couples choices that compete for a common capacity. Balancing may operate on an internal representation, while $\mathbf P$ contains only actual edge weights; unit row and column sums are therefore not a universal requirement on $\mathbf P$. This distinction allows nodes to remain unselected when the problem permits it. The case studies specify the internal representation and normalization rules.

Lower temperature amplifies score differences in the exponential weights, which can sharpen the selection but also reduce gradients when outputs saturate. It controls concentration and smoothness, not which selection is correct.

Refinement starts from the differentiable initialization $\mathcal I(u)$, such as a Softmax transformation of the baseline scores; this initial selection need not satisfy all constraints. Each round recomputes the relations in~\eqref{eq:layer-decision-relations}, scores the edges, and applies $\mathcal B_\tau$, defining the update $\mathcal U_\omega$ in~\eqref{eq:layer-overall-computation}. The changed direct edge weights and relation features let the shared scorer reconsider each edge after the preceding update, while the input attributes and node representations are reused.

All rounds and differentiable preprocessing remain on the computation graph. Freezing $\omega$ in Stage 2 preserves gradients with respect to the supplied input, including paths through any intermediate quantities used by the loss. This gives the complete input derivative required in~\eqref{eq:gradient}, without hard selection or decoder calls.

These operations encourage concentrated and structurally consistent selections, but refinement is not a descent algorithm and need not improve the objective at every round. Elementary output constraints also do not ensure full combinatorial feasibility. The case studies assess both the continuous representation and the executed decisions, following the evaluation principles in Section~\ref{sec:framework}.

\section{Case studies}
\label{sec:experiments}

This section evaluates the proposed framework on two edge-selection problems. The asymmetric traveling salesperson problem (ATSP) considers uncertainty in the objective, while the endpoint-free orienteering problem (EFOP) considers uncertainty in the constraints or in both the objective and constraints. For each problem, we describe how the framework is instantiated and compare its decision quality and computational cost with baseline methods. We then examine the effects of data characteristics and training-set size.

\subsection{Experimental setup}
\label{subsec:experimental-setup}

We generate synthetic contextual data using a polynomial construction adopted in previous predict-then-optimize studies \citep{elmachtoub2022smart,tang2024cave,mandi2025odece,tang2024pyepo,gupta2024directional}. Specifically, we adapt the TSP cost generator of PyEPO \citep{tang2024pyepo} to directed edges and use the same cost-generation mechanism for ATSP and EFOP.

For each instance $s$, we independently sample a contextual vector $z^{(s)}\sim\mathcal{N}(0,I_p)$. Each directed edge $(i,j)$ has a fixed coefficient vector $b_{ij}\in\mathbb{R}^p$, whose entries are independently generated as the product of a Bernoulli$(0.5)$ variable and a uniform variable on $[-2,2]$. The edge cost is
\begin{equation}
    c_{ij}^{(s)}
    =
    \ell_{ij}
    +
    \frac{\left(3+b_{ij}^{\top}z^{(s)}/\sqrt{p}\right)^q}
         {3^{q-1}}
    \varepsilon_{ij}^{(s)},
    \qquad i\neq j,
    \label{eq:synthetic-cost}
\end{equation}
where $\ell_{ij}$ is the Euclidean distance between nodes $i$ and $j$, $q$ is the polynomial degree, and $\varepsilon_{ij}^{(s)}\sim\mathcal{U}(1-\eta_{\mathrm{noise}},1+\eta_{\mathrm{noise}})$ is independent multiplicative noise. Half of the node coordinates are sampled uniformly from $[-2,2]^2$, and the remainder from $\mathcal{N}(0,I_2)$. The coordinates and coefficient vectors are sampled once for each dataset and held fixed across its training, validation, and test instances. Reverse edges have independently sampled coefficient vectors and noise, so their costs are generally different. Costs are rounded to four decimal places, and diagonal entries are set to zero. Additional problem-specific parameters are described in the corresponding case study.

The polynomial degree controls the nonlinearity of the contextual mapping, while $\eta_{\mathrm{noise}}$ controls the noise magnitude. For the linear predictors used here, degrees greater than one introduce model misspecification, allowing us to examine whether decision-focused training is beneficial when the predictor cannot exactly represent the underlying relationship. Changing the degree also changes the cost distribution, so it is not interpreted as an isolated measure of optimization difficulty. Both case studies consider $n\in\{50,100,200\}$, $p=10$, $\eta_{\mathrm{noise}}=0.5$, and $q\in\{4,6\}$ in the main experiments.

For each setting, training, validation, and test sets are disjoint, with the same split shared by all compared methods. ATSP uses 1,000 training instances per setting, whereas EFOP uses 2,000; both use 64 validation instances and 500 test instances. Each training run starts from a fresh model and optimizer state, without reusing checkpoints from previous experiments. We use model seed 0.

All methods use a linear prediction model with an intercept, $\hat w_{\theta}(z)=Wz+b$, where $z\in\mathbb{R}^p$ is the contextual feature vector and $\hat w_{\theta}(z)\in\mathbb{R}^{d_w}$ contains the predicted unknown parameters \citep{elmachtoub2022smart,berthet2020learning,tang2024cave,gupta2024directional,mandi2025odece}. This shared architecture controls for predictive capacity and allows us to evaluate decision quality under model misspecification induced by the nonlinear data-generating process.

The MSE baseline estimates $W$ and $b$ by ordinary least squares. Each decision-focused predictor is initialized from an independently computed least-squares fit on the same training data and subsequently updated using its respective objective. This initialization uses observed parameter values but no optimal decisions. Methods requiring solution labels or optimization oracles retain these components, as specified in each case study.

For DFLNOL, the optimization layer is trained on realized parameters without solution labels and then frozen while the predictor is trained. Both stages use the same training instances. In both problems, the NOL represents each node by a vector of 16 learned features. The edge-scoring MLP has one hidden layer of 32 units and produces a scalar score for each edge. The same scoring network is reused over four refinement rounds, each containing 50 Sinkhorn iterations. The initial edge-selection matrix is computed using a row-wise Softmax with temperature $0.05$ for both problems. Each refinement round then applies Sinkhorn normalization with temperature $0.1$.

During NOL training, constraint-penalty weights are updated dynamically according to the observed constraint violations, following Section~\ref{sec:module}. This weighting strategy is used for both ATSP and EFOP. Both DFLNOL training stages use Adam with batch size 32, with learning rates of $0.003$ for the NOL and $0.001s_{\mathrm{tr}}$ for the predictor, where $s_{\mathrm{tr}}$ is twice the mean off-diagonal realized cost over the corresponding training set. Checkpoints are evaluated on decoded validation decisions under the realized parameters: we minimize mean tour cost for ATSP and maximize mean executed reward for EFOP. We retain the best checkpoint and use a patience of five epochs for early stopping. Model selection requires neither optimal solutions nor optimal objective values. Decoding and any execution correction are used for evaluation, not gradient computation.

We evaluate decision quality using relative regret under the realized parameters. For test instance $s$, let $C_s$ denote the cost of the ATSP tour produced by a method and $R_s$ the reward collected in EFOP under its execution rule. Let $C_s^{\mathrm{ref}}$ and $R_s^{\mathrm{ref}}$ denote the corresponding reference values obtained by solving the instance with the realized parameters. Relative regret is computed as $(C_s-C_s^{\mathrm{ref}})/C_s^{\mathrm{ref}}$ for ATSP and $(R_s^{\mathrm{ref}}-R_s)/R_s^{\mathrm{ref}}$ for EFOP, both expressed as percentages. We report the mean and median relative regret to summarize overall decision quality, together with the 95th percentile (P95) and maximum observed regret to assess tail performance.

Training time is measured as wall-clock time for a complete run, including initialization, parameter fitting, validation, and model selection. It also includes required preprocessing, such as generating training solution labels, preparing oracle solutions, or pretraining a NOL. Both stages are included for DFLNOL. Dataset generation, test-reference computation, and hyperparameter searches are excluded. Inference time is the mean time required to produce a discrete decision for one test instance, measured with batch size one after warm-up. It includes prediction and, as applicable, neural optimization and decoding or Gurobi~\citep{gurobi} model construction and solution.

Experiments are implemented in Python 3.11 using PyTorch 2.14 and Gurobi 13.0.3 on a high-performance computing cluster. Training and solver jobs are allocated eight CPU cores and 64 GB of RAM on AMD EPYC 7663 nodes, with an NVIDIA A100 GPU with 80 GB of VRAM for GPU-based neural computations.

\subsection{Asymmetric traveling salesperson problem}
\label{subsec:atsp}

\subsubsection{Framework instantiation}
\label{subsubsec:atsp-instantiation}

The ATSP models applications such as courier delivery, robotic inspection, and cyclic production scheduling, where all required locations or tasks must be visited in a closed sequence and travel or setup costs depend on the direction of each transition. The problem seeks a minimum-cost directed tour that visits every node exactly once and returns to its starting node. On a complete directed graph without self-loops, it minimizes $\sum_{i\neq j}c_{ij}x_{ij}$ subject to unit incoming and outgoing degrees and the exclusion of subtours. The problem is NP-hard \citep{roberti2012models}. In our contextual setting, only edge costs are uncertain, so prediction errors affect tour quality but not feasibility.

The NOL maps a cost matrix $D\in\mathbb{R}^{n\times n}$ to a continuous edge-selection matrix $\mathbf{P}\in[0,1]^{n\times n}$, where $P_{ij}$ is the selection weight of directed edge $(i,j)$. Its input is the realized matrix $C$ during NOL pretraining and the predicted matrix $\hat C_{\theta}(z)$ during predictor training and deployment. Contextual features enter the predictor, not the NOL. The continuous output supports differentiation during training and is converted into a feasible tour by a separate decoder.

We normalize the input as $\bar D_{ij}=D_{ij}/s(D)$, with $s(D)=\frac{2}{n(n-1)}\sum_{i\neq j}\sqrt{D_{ij}^{2}+\epsilon_{\mathrm{norm}}^{2}}$, where $\epsilon_{\mathrm{norm}}>0$ ensures a positive scale. This differentiable normalization accommodates signed predictions and varying cost magnitudes without changing the ranking of tours under $D$.

The mean outgoing normalized cost initializes each node's feature, which the CAC encoder maps to a 16-dimensional representation. For each edge $(i,j)$, the scorer combines the two endpoint representations with five values: its normalized cost $\bar D_{ij}$, the current forward and reverse selection weights $P_{ij}$ and $P_{ji}$, and the structural features $(\mathbf{P}^{2})_{ij}$ and $(\mathbf{P}^{\top}\mathbf{P})_{ij}$. Here $(\mathbf{P}^{2})_{ij}=\sum_kP_{ik}P_{kj}$ measures two-step connection strength, while $(\mathbf{P}^{\top}\mathbf{P})_{ij}=\sum_kP_{ki}P_{kj}$ measures selection weight received from common predecessors. These features expose the surrounding selection pattern to the scorer. Together, the endpoint representations and five values form a 37-dimensional input.

A row-wise Softmax of negative normalized costs initializes $\mathbf{P}$. The shared scorer then updates the selection through four refinement rounds, each followed by Sinkhorn normalization. The diagonal is masked throughout, and no symmetry is imposed. Sinkhorn encourages unit row and column sums, but finite iterations can leave degree errors, and even exact balance permits fractional selections and disjoint cycles.

To address these limitations, NOL pretraining combines the soft cost with three structural terms:
\begin{equation}
    \mathcal{L}_{\mathrm{NOL}}
    =
    \frac{1}{s_{\mathrm{tr}}}\sum_{i\neq j}c_{ij}P_{ij}
    +\lambda_{\mathrm{deg}}\mathcal{L}_{\mathrm{deg}}
    +\beta_{\mathrm{con}}\mathcal{L}_{\mathrm{con}}
    +\lambda_{\mathrm{bin}}\mathcal{L}_{\mathrm{bin}}.
    \label{eq:atsp-nol-loss}
\end{equation}
The first term favors inexpensive edges and equals the scaled tour cost when $\mathbf{P}$ represents a Hamiltonian tour. The fixed training-set scale $s_{\mathrm{tr}}$ controls the loss magnitude and is distinct from the instance-dependent input scale $s(D)$. The degree and binarization multipliers follow Section~\ref{sec:module}, while $\beta_{\mathrm{con}}$ is fixed. Additive tolerance terms independent of the network parameters are omitted.

The degree penalty,
\begin{equation}
    \mathcal{L}_{\mathrm{deg}}
    =
    \sum_i\left(\sum_{j\neq i}P_{ij}-1\right)^2
    +
    \sum_j\left(\sum_{i\neq j}P_{ij}-1\right)^2,
\end{equation}
penalizes outgoing and incoming degree errors separately. It complements Sinkhorn normalization by discouraging residual imbalance after a finite number of iterations.

Degree consistency alone does not exclude subtours. We therefore use
\begin{equation}
    \mathcal{L}_{\mathrm{con}}
    =
    -\log\left|
    \det\left(
        \operatorname{Diag}(\mathbf{P}\mathbf{1})-\mathbf{P}
        +\frac{\mathbf{1}\mathbf{1}^{\top}}{n}
        +\epsilon_{\mathrm{con}}I
    \right)\right|.
\end{equation}
The first two terms form a directed graph Laplacian, and the rank-one term removes its zero eigenvalue associated with the constant vector. For a binary cycle cover, additional disconnected cycles introduce further zero eigenvalues. With the small stabilizer $\epsilon_{\mathrm{con}}>0$, these contribute small factors to the determinant and increase the penalty. This discourages disconnected cycles but is not an exact subtour-elimination constraint.

The binarization penalty is $\mathcal{L}_{\mathrm{bin}}=\sum_{i\neq j}P_{ij}(1-P_{ij})$. Each summand vanishes at 0 and 1 and is positive between them. Combined with the degree conditions, it encourages selection mass to concentrate on individual edges, making the continuous representation closer to a discrete decision. The penalties guide the output structure; feasibility is enforced by decoding.

During predictor training, the NOL is frozen and the loss is $\sum_{i\neq j}c_{ij}(\mathbf{P}_{\theta})_{ij}/(n s_{\mathrm{tr}})$, where $\mathbf{P}_{\theta}$ is induced by $\hat C_{\theta}(z)$. Thus, predictions are evaluated through the continuous output under realized costs, without differentiating through the decoder.

Decoding first constructs a directed cycle cover by solving the assignment problem
\begin{equation}
    \max_{\substack{\sigma\in\boldsymbol{\sigma}_n\\
                    \sigma(i)\neq i\;\forall i}}
    \sum_i\log\bigl(\max\{P_{i,\sigma(i)},\delta\}\bigr),
    \label{eq:atsp-assignment}
\end{equation}
where $\boldsymbol{\sigma}_n$ is the set of node permutations and $\delta>0$ safeguards the logarithm. Each node then has exactly one successor and one predecessor.

If multiple cycles remain, we merge two by replacing $(a,\sigma(a))$ and $(b,\sigma(b))$ with $(a,\sigma(b))$ and $(b,\sigma(a))$. At each step, we choose nodes in different cycles that minimize the predicted-cost change $\Delta_{ab}=\hat c_{a,\sigma(b)}+\hat c_{b,\sigma(a)}-\hat c_{a,\sigma(a)}-\hat c_{b,\sigma(b)}$. Each merge preserves the degree conditions and reduces the cycle count by one, yielding a Hamiltonian tour on the complete graph.

\subsubsection{Experimental evaluation}
\label{subsubsec:atsp-experiment}

We compare against MSE, SPO+ \citep{elmachtoub2022smart}, CaVE \citep{tang2024cave}, PGC \citep{gupta2024directional}, and PFYL \citep{berthet2020learning}. MSE is the standard prediction-focused baseline. CaVE uses optimal training solutions to guide cost prediction through cone alignment. PGC uses central differences of optimal objective values, while PFYL learns from optimal training solutions and solutions under perturbed predicted costs. Methods retain their required labels and oracle calls, whose costs are included in training time.

We evaluate SPO+ with either its standard Gurobi oracle or a neural-oracle variant introduced in this study. SPO+ requires an optimization oracle that maps a supplied cost vector to a discrete tour for computing predictor updates. In our variant, a pretrained NOL followed by decoding replaces this oracle, both when constructing reference tours under the observed costs and when answering shifted-cost queries during training. The NOL remains frozen throughout predictor training. This variant examines whether our learned optimization layer can accelerate an existing DFL method beyond its differentiable use in DFLNOL. Since the neural oracle returns heuristic tours, the exact-oracle guarantees of SPO+ do not directly apply.

Our method is evaluated as DFLNOL, which deploys the NOL with assignment-and-merging decoding, and DFLNOL $\rightarrow$ Gurobi, which instead solves the predicted ATSP with Gurobi. All baseline methods also use Gurobi at deployment, including SPO+ with the neural training oracle.

Table~\ref{tab:atsp_results} compares decision quality and computational cost across the six ATSP settings. Solver-based deployment assesses the learned predictions under a common optimization procedure, while DFLNOL additionally evaluates the quality--efficiency trade-off of neural deployment.

\begin{table*}[H]
\centering
\caption{ATSP results. Boldface indicates the best performance among decision-focused methods.}
\label{tab:atsp_results}
\small
\setlength{\tabcolsep}{7pt}
\renewcommand{\arraystretch}{1.0}

\resizebox{\textwidth}{!}{%
\begin{tabular}{@{}cclrrrrr@{}r@{}rr@{}}
\toprule
\multirow{2}{*}{Nodes}
& \multirow{2}{*}{Degree}
& \multirow{2}{*}{Method}
& \multicolumn{4}{c}{Regret (\%)}
& \multicolumn{3}{c}{\multirow{2}{*}{\shortstack{Training time\\(h:m:s)}}}
& \multirow{2}{*}{\shortstack{Inference time\\(ms/instance)}} \\
\cmidrule(lr){4-7}
& & & Mean & Median & P95 & Worst
& \multicolumn{3}{c}{} & \\
\midrule

\multirow{16}{*}{50}
& \multirow{8}{*}{4}
& MSE
& 33.19 & 31.33 & 55.48 & 123.20
& \skinf & \skinf & \skzer0.14 & 86.69 \\
& & SPO+ 
& 9.76 & 9.45 & 17.00 & 37.34
& \skinf & 6: & 50.91 & 99.81 \\
& & SPO+ (NOL oracle)
& 12.88 & 11.67 & 23.93 & 48.74
& \skinf & \skinf & 22.52 & 108.07 \\
& & CaVE
& 10.25 & 9.80 & 17.59 & 38.10
& \skinf & 1: & 23.70 & 91.32 \\
& & PGC
& 12.56 & 11.68 & 22.44 & 37.12
& \skinf & 12: & 52.79 & 97.36 \\
& & PFYL
& 9.26 & 8.78 & 16.09 & \textbf{29.62}
& \skinf & 12: & 43.95 & 92.15 \\
& & DFLNOL 
& 11.62 & 10.89 & 20.58 & 39.39
& \skinf & \skinf & \textbf{17.31} & \textbf{9.38} \\
& & DFLNOL $\to$ Gurobi
& \textbf{7.97} & \textbf{7.65} & \textbf{14.33} & 39.98
& \skinf & \skinf & \textbf{17.31} & 118.51 \\
\cmidrule(lr){2-11}

& \multirow{8}{*}{6}
& MSE
& 59.65 & 58.22 & 90.86 & 148.90
& \skinf & \skinf & \skzer0.08 & 82.31 \\
& & SPO+ 
& 14.35 & 13.10 & 24.97 & 49.57
& \skinf & 9: & 02.13 & 105.01 \\
& & SPO+ (NOL oracle)
& 18.78 & 17.40 & 36.75 & 62.87
& \skinf & \skinf & 32.74 & 99.71 \\
& & CaVE
& 12.42 & 11.50 & 22.58 & 65.87
& \skinf & 1: & 16.97 & 100.58 \\
& & PGC
& 26.39 & 25.18 & 46.04 & 80.92
& \skinf & 24: & 23.61 & 78.36 \\
& & PFYL
& 13.42 & 12.43 & 23.69 & 77.33
& \skinf & 15: & 42.14 & 101.27 \\
& & DFLNOL 
& 15.46 & 13.59 & 27.93 & 41.51
& \skinf & \skinf & \textbf{26.56} & \textbf{9.20} \\
& & DFLNOL $\to$ Gurobi
& \textbf{10.01} & \textbf{9.45} & \textbf{18.52} & \textbf{36.53}
& \skinf & \skinf & \textbf{26.56} & 115.54 \\
\midrule

\multirow{16}{*}{100}
& \multirow{8}{*}{4}
& MSE
& 64.81 & 62.73 & 111.32 & 153.93
& \skinf & \skinf & \skzer0.35 & 356.73 \\
& & SPO+ 
& 13.19 & 11.65 & 24.24 & 42.89
& \skinf & 25: & 23.53 & 372.05 \\
& & SPO+ (NOL oracle)
& 16.36 & 14.49 & 28.95 & 47.63
& \skinf & \skinf & 58.84 & 365.61 \\
& & CaVE
& 12.50 & 12.24 & 18.78 & 28.40
& \skinf & 5: & 45.28 & 375.12 \\
& & PGC
& 18.28 & 16.41 & 30.58 & 51.43
& \skinf & 57: & 55.36 & 447.50 \\
& & PFYL
& 11.78 & 11.13 & 19.26 & 27.08
& \skinf & 36: & 51.84 & 365.07 \\
& & DFLNOL 
& 13.58 & 12.91 & 22.81 & 37.35
& \skinf & \skinf & \textbf{25.45} & \textbf{12.04} \\
& & DFLNOL $\to$ Gurobi
& \textbf{9.60} & \textbf{9.26} & \textbf{15.28} & \textbf{24.09}
& \skinf & \skinf & \textbf{25.45} & 470.70 \\
\cmidrule(lr){2-11}

& \multirow{8}{*}{6}
& MSE
& 121.79 & 121.96 & 176.80 & 214.40
& \skinf & \skinf & \skzer0.36 & 443.81 \\
& & SPO+ 
& 19.17 & 16.92 & 34.72 & 52.75
& 1: & 23: & 28.58 & 374.63 \\
& & SPO+ (NOL oracle)
& 26.84 & 23.83 & 50.12 & 79.28
& \skinf & 1: & 39.59 & 345.70 \\
& & CaVE
& 32.25 & 28.76 & 59.70 & 104.39
& \skinf & 5: & 37.89 & 308.13 \\
& & PGC
& 47.11 & 44.70 & 77.92 & 112.89
& 1: & 50: & 06.58 & 339.93 \\
& & PFYL
& 15.04 & 14.16 & 24.56 & 36.46
& 1: & 34: & 27.38 & 578.68 \\
& & DFLNOL 
& 19.86 & 17.94 & 36.35 & 61.36
& \skinf & \skinf & \textbf{36.01} & \textbf{12.10} \\
& & DFLNOL $\to$ Gurobi
& \textbf{12.97} & \textbf{11.93} & \textbf{24.01} & \textbf{35.06}
& \skinf & \skinf & \textbf{36.01} & 461.42 \\
\midrule

\multirow{16}{*}{200}
& \multirow{8}{*}{4}
& MSE
& 97.84 & 92.63 & 162.17 & 298.53
& \skinf & \skinf & \skzer1.31 & 1255.47 \\
& & SPO+ 
& 16.27 & 13.98 & 31.34 & 114.33
& 4: & 17: & 06.06 & 1690.21 \\
& & SPO+ (NOL oracle)
& 20.01 & 16.60 & 38.19 & 119.33
& \skinf & 5: & 25.72 & 2798.22 \\
& & CaVE
& 13.92 & 13.69 & 18.81 & \textbf{31.36}
& \skinf & 33: & 05.28 & 1858.16 \\
& & PGC
& 26.41 & 23.10 & 49.49 & 121.12
& 4: & 45: & 28.31 & 1366.24 \\
& & PFYL
& 12.95 & 12.21 & 19.21 & 53.97
& 3: & 28: & 07.39 & 1800.27 \\
& & DFLNOL 
& 13.99 & 13.08 & 21.71 & 58.10
& \skinf & \textbf{1:} & \textbf{01.27} & \textbf{14.21} \\
& & DFLNOL $\to$ Gurobi
& \textbf{10.11} & \textbf{9.46} & \textbf{16.74} & 50.36
& \skinf & \textbf{1:} & \textbf{01.27} & 2096.04 \\
\cmidrule(lr){2-11}

& \multirow{8}{*}{6}
& MSE
& 178.38 & 176.43 & 252.26 & 356.62
& \skinf & \skinf & \skzer1.36 & 1223.35 \\
& & SPO+ 
& 25.65 & 21.43 & 52.77 & 115.47
& 12: & 57: & 44.75 & 1721.61 \\
& & SPO+ (NOL oracle)
& 35.17 & 29.71 & 70.56 & 168.13
& \skinf & 7: & 58.69 & 2492.82 \\
& & CaVE
& 61.19 & 55.64 & 114.48 & 189.28
& \skinf & 37: & 12.62 & 1249.67 \\
& & PGC
& 90.27 & 88.27 & 138.40 & 233.38
& 8: & 36: & 29.07 & 1212.22 \\
& & PFYL
& 16.41 & 15.84 & \textbf{24.56} & \textbf{43.80}
& 11: & 03: & 53.50 & 2381.03 \\
& & DFLNOL 
& 21.05 & 19.21 & 37.01 & 82.44
& \skinf & \textbf{1:} & \textbf{34.41} & \textbf{14.49} \\
& & DFLNOL $\to$ Gurobi
& \textbf{14.06} & \textbf{12.51} & 25.51 & 63.98
& \skinf & \textbf{1:} & \textbf{34.41} & 2127.08 \\
\bottomrule
\end{tabular}%
}
\end{table*}

Decision-focused training substantially reduces mean regret relative to MSE in all six settings. Without Gurobi deployment, DFLNOL improves both average and upper-tail performance. For example, at 200 nodes and degree 6, mean regret decreases from $178.38\%$ to $21.05\%$, while P95 decreases from $252.26\%$ to $37.01\%$. Its advantage over MSE is maintained as the problem size increases: at degree 4, DFLNOL's mean regret rises modestly from $11.62\%$ at 50 nodes to $13.99\%$ at 200 nodes, whereas MSE's increases from $33.19\%$ to $97.84\%$.

Neural deployment also remains competitive with solver-based DFL. At 200 nodes, DFLNOL achieves lower mean, median, P95, and worst-case regret than SPO+ with a Gurobi oracle for both polynomial degrees, despite requiring no ATSP solver at deployment. PFYL nevertheless achieves lower mean regret in all six settings, so neural deployment does not uniformly outperform solver-based alternatives. With Gurobi deployment, DFLNOL achieves the lowest mean and median regret in every setting and the lowest P95 in five of the six settings. These results indicate that training through the continuous NOL output produces predictions that remain effective under a separate optimization solver, without requiring solver-generated solution labels or ATSP solver calls during either training stage.

DFLNOL has the shortest training time among the decision-focused methods in all six settings, including the cost of NOL pretraining. Even for 200-node instances, training finishes in under two minutes, whereas the competing methods require several minutes to hours. MSE is cheaper to fit but produces substantially higher regret. Neural deployment is also the fastest in every setting, taking less than 15 milliseconds per instance. For 200-node instances, Gurobi-based deployment takes more than one second per instance. The two DFLNOL variants therefore offer a practical trade-off: neural deployment provides fast decisions, while Gurobi deployment achieves lower mean regret at greater inference cost.

The NOL can also accelerate an existing DFL method by replacing its optimization oracle. In SPO+, the pretrained NOL and decoder provide the discrete tours required for predictor updates without repeated Gurobi calls. At 200 nodes and degree 6, this replacement reduces training time from approximately 13 hours to 8 minutes, although mean regret increases from $25.65\%$ to $35.17\%$. This result suggests that the NOL can serve as a computationally inexpensive approximate oracle within an existing DFL training procedure, with a trade-off between training efficiency and decision quality.

\subsection{Endpoint-free orienteering problem}

\subsubsection{Framework instantiation}
\label{subsubsec:efop-instantiation}

The EFOP models applications such as sightseeing itinerary planning, mobile sensing, and selective inspection, where a traveler or agent starts from a given location and selects sites to maximize the collected reward within a travel budget. The problem seeks a maximum-reward directed path starting at a designated node 0, subject to a travel budget $B$. Each other node may be visited at most once, the endpoint is unrestricted, and no return to the start is required. Let $V^+=V\setminus\{0\}$ denote the reward-bearing nodes, with $r_0=0$. We consider uncertainty in travel costs alone or in both costs and rewards. The budget is known when planning, but uncertain travel costs can make a planned path infeasible under the realized parameters. 

The NOL takes a cost matrix, a reward vector, and the budget $B$ as inputs and outputs a continuous edge-selection matrix $\mathbf{P}$. During NOL pretraining, the inputs are the realized parameters $(C,r,B)$. During predictor training and deployment, they are $(\hat C_\theta(z),r,B)$ for cost-only prediction and $(\hat C_\theta(z),\hat r_\theta(z),B)$ for joint prediction. Contextual features $z$ enter the predictor rather than the NOL.

Costs and rewards are normalized by twice their respective instance means, with positive numerical floors. The normalized NOL inputs are denoted by $\bar c_{ij}$ and $\bar r_i$, using realized or predicted parameters as appropriate to each stage. The cost mean excludes self-loops and edges returning to node 0, while the reward mean excludes node 0. Each node has four initial features: normalized reward, mean outgoing normalized cost, normalized budget $B/(n s_C)$, and a start-node indicator, where $s_C$ is the cost scale of the current input. The CAC encoder maps these features to node representations. For edge $(i,j)$, the scorer combines its two endpoint representations with $\bar c_{ij}$, $P_{ij}$, $P_{ji}$, $(\mathbf{P}^{2})_{ij}$, and $(\mathbf{P}^{\top}\mathbf{P})_{ij}$. These describe the edge cost, forward and reverse selection weights, two-step connections, and shared predecessors, respectively.

A path involves three choices: traveling to another node, leaving a node unvisited, and terminating. We encode these choices in an internal matrix $\mathbf{Q}$. An entry $Q_{ij}$ with $j\in V^+$ and $i\neq j$ represents travel; $Q_{ii}$ for $i\in V^+$ represents skipping node $i$; and $Q_{i0}$ represents termination at node $i$. The entry $Q_{00}$ permits an empty path. Termination entries are virtual and incur no return cost. The actual travel matrix has $P_{ij}=Q_{ij}$ for $j\in V^+$ and $i\neq j$, and $P_{ij}=0$ otherwise. Thus, $y_j=\sum_iP_{ij}$ measures the soft visitation of customer $j$, rather than requiring every customer to be visited.

The network performs four shared refinement rounds. Actual travel edges receive energies $\bar c_{ij}+0.1\tanh(f_\omega)$ from the scorer. Skipping node $i$ has energy $\kappa_{\mathrm{skip}}\bar r_i$, where $\kappa_{\mathrm{skip}}>0$ is learned, making it more costly to skip a high-reward node; termination has zero energy. Each round applies Sinkhorn normalization to $\mathbf{Q}$ and extracts $\mathbf{P}$ for the next update. Balancing $\mathbf{Q}$ coordinates travel, skipping, and termination, giving $y_j\approx1-Q_{jj}$.

For NOL pretraining, define the soft reward as $R(\mathbf{P};r)=\sum_{j\in V^+}r_jy_j$ and travel cost as $T(\mathbf{P};C)=\sum_{i,j}c_{ij}P_{ij}$. The training loss is
\begin{equation}
    \mathcal{L}_{\mathrm{NOL}}
    =
    -\frac{R(\mathbf{P};r)}{\sum_{j\in V^+}r_j}
    +\lambda_{\mathrm{bud}}\mathcal{L}_{\mathrm{bud}}
    +\lambda_{\mathrm{bal}}\mathcal{L}_{\mathrm{bal}}
    +\beta_{\mathrm{con}}\mathcal{L}_{\mathrm{con}},
    \label{eq:efop-nol-loss}
\end{equation}
where additive tolerance terms independent of the network parameters are omitted. The budget and balance multipliers follow the dynamic update rule in Section~\ref{sec:module}, while $\beta_{\mathrm{con}}$ is a fixed structural regularization coefficient.

The negative reward term encourages valuable visits. The budget penalty $\mathcal{L}_{\mathrm{bud}}=[T(\mathbf{P};C)/B-1]_+$ penalizes expenditure above the budget, where $[a]_+=\max\{a,0\}$. The balance penalty is $\mathcal{L}_{\mathrm{bal}}=\frac{1}{n}\left(\|\mathbf{Q}\mathbf{1}-\mathbf{1}\|_2^2+\|\mathbf{Q}^{\top}\mathbf{1}-\mathbf{1}\|_2^2\right)$. This discourages residual imbalance after finite Sinkhorn iterations while retaining the option to skip nodes and terminate.

Balanced selections may still contain disconnected cycles whose rewards cannot be collected from the start. Let $\mathbf{P}_{+}$ be the customer-to-customer submatrix. We use
\begin{equation}
    \begin{aligned}
        \mathcal{L}_{\mathrm{con}}
        ={}&(n-1)\log(1+\epsilon_{\mathrm{con}})\\
        &-\log\det\left((1+\epsilon_{\mathrm{con}})I-\mathbf{P}_{+}\right).
    \end{aligned}
    \label{eq:efop-connectivity}
\end{equation}
This term is zero for a binary path because its customer subgraph is acyclic. A disconnected directed cycle introduces an eigenvalue of one, reducing the stabilized determinant and increasing the penalty. The regularizer discourages cycles without requiring skipped nodes to be connected, but does not guarantee path feasibility.

During predictor training, the NOL is frozen, but gradients pass through its continuous output to the predictor. The soft visitation $y_j=\sum_iP_{ij}$ measures incoming selection mass, which does not necessarily indicate that node $j$ is reachable from the start. For example, an isolated cycle can assign full visitation to its nodes without connecting them to node 0. To reduce reward attributed to such selections, we replace $y_j$ in the reward calculation by
\begin{equation}
    \tilde y_j
    =
    \bigl[
        \min\{y_j,\ a_0,\ \min\nolimits_{i\in V^+\setminus\{j\}} a_{ij}^{\mathrm{ext}}\}
    \bigr]_+.
    \label{eq:efop-reward-credit}
\end{equation}
Here $a_0=\sum_{k\in V^+}P_{0k}$ is the mass leaving the start, and $a_{ij}^{\mathrm{ext}}=y_i+y_j-P_{ij}-P_{ji}$ is the total mass entering the node pair $\{i,j\}$ from outside that pair. Thus, a node receives no visitation credit if no mass leaves the start, and an isolated two-node cycle receives no credit because its external incoming mass is zero. The correction only reduces the reward attributed to the output; it does not modify $\mathbf{P}$. These local checks leave valid binary paths unchanged but do not establish connectivity for every continuous selection.

Predicted costs may also underestimate the expenditure required to collect reward. We therefore evaluate the soft travel cost under the realized matrix $C$ and define its budget ratio as $\rho=T(\mathbf{P}_\theta;C)/B$. The predictor minimizes
\begin{equation}
    \mathcal{L}_{\mathrm{DFL}}
    =
    -\sum_{j\in V^+}r_j\tilde y_j
    +s_R\left(
        \gamma[\rho-1]_+^2+\eta_{\mathrm{cost}}\rho
    \right),
    \label{eq:efop-dfl-loss}
\end{equation}
where $s_R$ is the mean total reward in the training set. The first term encourages reward collection after the connectivity correction. The squared penalty is zero within budget and increases with budget excess. The linear term introduces an additional preference for lower expenditure, including below the budget, and therefore trades some reward against travel cost. We use $\gamma=1$ for cost-only prediction, $\gamma=3$ for joint prediction, and $\eta_{\mathrm{cost}}=0.25$. Costs and rewards in this loss are realized values, while the continuous selection is induced by the predicted parameters. Training differentiates through this piecewise-differentiable objective and the frozen NOL, without using the hard decoder.

For neural deployment, a reward-aware greedy decoder starts at node 0. At each step, it selects the unvisited node affordable under the remaining predicted budget with the largest score
$
    s_{ij}=P_{ij}\left(\frac{\hat r_j}{\hat c_{ij}}\right).
$
For cost-only prediction, $\hat r_j=r_j$ because rewards are known. The score combines the NOL's edge preference $P_{ij}$ with the reward per unit predicted cost $\hat r_j/\hat c_{ij}$. Decoding stops when no affordable node remains, without a return edge or local search.

All methods are evaluated using the same execution rule. The planned node order is fixed, and collected reward is computed from the longest prefix whose cumulative realized travel cost does not exceed $B$. A node earns no reward if it cannot be reached within budget. No reordering or replanning is allowed, so evaluation measures realized reward rather than the reward of a potentially infeasible plan.

\subsubsection{Experimental evaluation}
\label{subsubsec:efop-experiment}

We compare against MSE and Odece \citep{mandi2025odece}. MSE predicts the unknown parameters by minimizing prediction error and then solves the predicted problem with Gurobi. Odece is designed for uncertain constraint parameters: its training objective encourages predictions that preserve the feasibility of reference optimal solutions while discouraging decisions that violate the realized constraints. We evaluate Odece in the cost-only setting, where rewards are known. The joint cost-and-reward setting compares MSE with the two DFLNOL deployment variants.

We also evaluate Odece with a neural training oracle. In this variant, a pretrained NOL followed by decoding supplies the discrete candidate paths used during predictor updates, replacing the corresponding Gurobi calls. The NOL remains frozen, and gradients are not propagated through these discrete paths. Reference optimal solutions are retained, and validation and deployment still use Gurobi. This variant tests whether the NOL can accelerate a constraint-focused DFL method without replacing its training objective.

DFLNOL uses the neural decoder for deployment, while DFLNOL $\rightarrow$ Gurobi solves the predicted problem with Gurobi. The two variants share the same trained predictor. All methods are evaluated by the reward collected under the realized-cost execution rule in Section~\ref{subsubsec:efop-instantiation}. Table~\ref{tab:op_results} reports results for both uncertainty settings across the three problem sizes and two polynomial degrees.

\begin{table*}[H]
\centering
\caption{EFOP results. Boldface indicates the best performance among decision-focused methods.}
\label{tab:op_results}
\small
\setlength{\tabcolsep}{7pt}
\renewcommand{\arraystretch}{0.92}

\resizebox{\textwidth}{!}{%
\begin{tabular}{@{}ccclrrrrr@{}r@{}rr@{}}
\toprule
\multirow{2}{*}{Nodes}
& \multirow{2}{*}{Degree}
& \multirow{2}{*}{\shortstack{Uncertain\\components}}
& \multirow{2}{*}{Method}
& \multicolumn{4}{c}{Regret (\%)}
& \multicolumn{3}{c}{\multirow{2}{*}{\shortstack{Training time\\(h:m:s)}}}
& \multirow{2}{*}{\shortstack{Inference time\\(ms/instance)}} \\
\cmidrule(lr){5-8}
& & & & Mean & Median & P95 & Worst
& \multicolumn{3}{c}{} & \\
\midrule

\multirow{16}{*}{50}
& \multirow{8}{*}{4}
& \multirow{5}{*}{Constraints}
& MSE
& 33.19 & 32.06 & 51.03 & 70.66
& \skinf & \skinf & \skzer0.08 & 251.55 \\
& & & Odece 
& 21.68 & 20.36 & 39.74 & 58.51
& \skinf & 28: & 18.94 & 428.33 \\
& & & Odece (NOL oracle)
& 23.14 & 22.34 & 38.18 & 51.91
& \skinf & 6: & 19.22 & 468.97 \\
& & & DFLNOL  
& \textbf{12.61} & \textbf{12.14} & \textbf{21.85} & \textbf{45.70}
& \skinf & \skinf & \textbf{21.77} & \textbf{9.74} \\
& & & DFLNOL $\to$ Gurobi
& 18.56 & 17.62 & 31.14 & 46.99
& \skinf & \skinf & \textbf{21.77} & 247.47 \\
\cmidrule(lr){3-12}

& & \multirow{3}{*}{\shortstack{Constraints\\and objective}}
& MSE
& 31.10 & 30.16 & 46.35 & 73.10
& \skinf & \skinf & \skzer0.09 & 235.91 \\
& & & DFLNOL  
& \textbf{12.78} & \textbf{12.01} & 23.55 & \textbf{47.71}
& \skinf & \skinf & \textbf{23.06} & \textbf{9.76} \\
& & & DFLNOL $\to$ Gurobi
& 12.87 & 12.33 & \textbf{22.10} & 49.73
& \skinf & \skinf & \textbf{23.06} & 199.49 \\
\cmidrule(lr){2-12}

& \multirow{8}{*}{6}
& \multirow{5}{*}{Constraints}
& MSE
& 46.64 & 45.93 & 69.60 & 92.75
& \skinf & \skinf & \skzer0.19 & 224.86 \\
& & & Odece 
& 28.32 & 26.41 & 53.30 & 99.37
& 1: & 05: & 43.24 & 556.95 \\
& & & Odece (NOL oracle)
& 30.52 & 29.33 & 52.14 & 82.87
& \skinf & 6: & 10.16 & 479.03 \\
& & & DFLNOL  
& \textbf{14.52} & \textbf{13.49} & \textbf{29.49} & \textbf{46.36}
& \skinf & \skinf & \textbf{41.68} & \textbf{9.59} \\
& & & DFLNOL $\to$ Gurobi
& 24.49 & 22.40 & 46.38 & 69.19
& \skinf & \skinf & \textbf{41.68} & 236.20 \\
\cmidrule(lr){3-12}

& & \multirow{3}{*}{\shortstack{Constraints\\and objective}}
& MSE
& 40.71 & 40.65 & 58.43 & 69.99
& \skinf & \skinf & \skzer0.19 & 211.14 \\
& & & DFLNOL  
& \textbf{12.48} & \textbf{11.27} & \textbf{24.36} & 48.06
& \skinf & \textbf{1:} & \textbf{15.13} & \textbf{9.82} \\
& & & DFLNOL $\to$ Gurobi
& 16.22 & 15.42 & 26.04 & \textbf{44.59}
& \skinf & \textbf{1:} & \textbf{15.13} & 162.15 \\
\midrule

\multirow{16}{*}{100}
& \multirow{8}{*}{4}
& \multirow{5}{*}{Constraints}
& MSE
& 44.66 & 43.77 & 62.56 & 81.58
& \skinf & \skinf & \skzer0.32 & 1047.76 \\
& & & Odece 
& 24.82 & 24.08 & 40.17 & 55.60
& 3: & 29: & 21.88 & 2522.82 \\
& & & Odece (NOL oracle)
& 26.26 & 25.68 & 40.04 & 53.82
& \skinf & 35: & 18.40 & 2577.08 \\
& & & DFLNOL  
& \textbf{15.00} & \textbf{14.32} & \textbf{23.70} & \textbf{37.49}
& \skinf & \skinf & \textbf{30.28} & \textbf{13.14} \\
& & & DFLNOL $\to$ Gurobi
& 20.04 & 18.59 & 33.86 & 50.61
& \skinf & \skinf & \textbf{30.28} & 1118.97 \\
\cmidrule(lr){3-12}

& & \multirow{3}{*}{\shortstack{Constraints\\and objective}}
& MSE
& 43.14 & 42.61 & 58.98 & 78.10
& \skinf & \skinf & \skzer0.32 & 1008.55 \\
& & & DFLNOL  
& \textbf{13.25} & \textbf{12.69} & \textbf{20.56} & 36.36
& \skinf & \skinf & \textbf{47.36} & \textbf{12.66} \\
& & & DFLNOL $\to$ Gurobi
& 19.99 & 20.08 & 25.65 & \textbf{32.81}
& \skinf & \skinf & \textbf{47.36} & 751.51 \\
\cmidrule(lr){2-12}

& \multirow{8}{*}{6}
& \multirow{5}{*}{Constraints}
& MSE
& 60.51 & 60.34 & 78.60 & 85.85
& \skinf & \skinf & \skzer0.36 & 1037.25 \\
& & & Odece 
& 34.34 & 32.26 & 58.90 & 77.69
& 8: & 13: & 12.20 & 4141.82 \\
& & & Odece (NOL oracle)
& 32.06 & 31.26 & 51.84 & 75.44
& \skinf & 40: & 08.30 & 2401.38 \\
& & & DFLNOL  
& \textbf{22.00} & \textbf{20.75} & \textbf{38.97} & \textbf{48.72}
& \skinf & \skinf & \textbf{50.48} & \textbf{12.47} \\
& & & DFLNOL $\to$ Gurobi
& 27.92 & 26.49 & 45.79 & 65.53
& \skinf & \skinf & \textbf{50.48} & 1031.86 \\
\cmidrule(lr){3-12}

& & \multirow{3}{*}{\shortstack{Constraints\\and objective}}
& MSE
& 56.81 & 56.69 & 73.06 & 88.30
& \skinf & \skinf & \skzer0.31 & 913.17 \\
& & & DFLNOL  
& \textbf{23.57} & 22.80 & \textbf{39.64} & \textbf{59.90}
& \skinf & \skinf & \textbf{45.07} & \textbf{12.64} \\
& & & DFLNOL $\to$ Gurobi
& 25.02 & \textbf{22.35} & 47.53 & 70.62
& \skinf & \skinf & \textbf{45.07} & 710.90 \\
\midrule

\multirow{16}{*}{200}
& \multirow{8}{*}{4}
& \multirow{5}{*}{Constraints}
& MSE
& 52.53 & 53.36 & 68.81 & 78.63
& \skinf & \skinf & \skzer1.25 & 5671.09 \\
& & & Odece 
& 30.38 & 29.80 & 47.41 & 66.96
& 20: & 54: & 12.98 & 14651.47 \\
& & & Odece (NOL oracle)
& 28.59 & 28.37 & 42.14 & 58.13
& 2: & 39: & 16.19 & 12483.19 \\
& & & DFLNOL  
& \textbf{14.83} & \textbf{14.26} & \textbf{21.82} & \textbf{31.09}
& \skinf & \textbf{1:} & \textbf{40.95} & \textbf{14.52} \\
& & & DFLNOL $\to$ Gurobi
& 21.51 & 20.85 & 31.04 & 43.79
& \skinf & \textbf{1:} & \textbf{40.95} & 6820.39 \\
\cmidrule(lr){3-12}

& & \multirow{3}{*}{\shortstack{Constraints\\and objective}}
& MSE
& 51.52 & 52.14 & 66.34 & 73.55
& \skinf & \skinf & \skzer1.26 & 5125.83 \\
& & & DFLNOL  
& \textbf{14.75} & \textbf{14.08} & \textbf{21.15} & 39.27
& \skinf & \textbf{1:} & \textbf{14.13} & \textbf{14.13} \\
& & & DFLNOL $\to$ Gurobi
& 19.38 & 19.36 & 25.18 & \textbf{28.56}
& \skinf & \textbf{1:} & \textbf{14.13} & 3811.73 \\
\cmidrule(lr){2-12}

& \multirow{8}{*}{6}
& \multirow{5}{*}{Constraints}
& MSE
& 68.05 & 68.45 & 81.12 & 88.65
& \skinf & \skinf & \skzer1.28 & 5433.74 \\
& & & Odece 
& 37.17 & 35.31 & 59.76 & 80.48
& 39: & 12: & 31.00 & 22287.56 \\
& & & Odece (NOL oracle)
& 36.64 & 36.25 & 54.03 & 68.86
& 3: & 25: & 31.70 & 24603.96 \\
& & & DFLNOL  
& \textbf{22.65} & \textbf{21.95} & \textbf{33.83} & \textbf{47.75}
& \skinf & \textbf{1:} & \textbf{51.38} & \textbf{14.54} \\
& & & DFLNOL $\to$ Gurobi
& 26.20 & 24.96 & 39.85 & 62.59
& \skinf & \textbf{1:} & \textbf{51.38} & 6188.14 \\
\cmidrule(lr){3-12}

& & \multirow{3}{*}{\shortstack{Constraints\\and objective}}
& MSE
& 64.65 & 64.97 & 77.50 & 85.98
& \skinf & \skinf & \skzer1.25 & 4455.52 \\
& & & DFLNOL  
& 27.76 & 26.45 & \textbf{45.14} & 66.99
& \skinf & \skinf & \textbf{54.83} & \textbf{14.18} \\
& & & DFLNOL $\to$ Gurobi
& \textbf{27.51} & \textbf{26.42} & 45.87 & \textbf{64.70}
& \skinf & \skinf & \textbf{54.83} & 4195.23 \\
\bottomrule
\end{tabular}%
}

\end{table*}

Both DFLNOL variants improve mean, median, P95, and worst-case regret relative to MSE in all twelve settings. The improvement therefore extends beyond average performance to the upper tail. For example, with 200 nodes, degree 4, and uncertain costs, neural deployment reduces mean regret from $52.53\%$ to $14.83\%$ and P95 from $68.81\%$ to $21.82\%$. The Gurobi-deployed variant also improves on MSE under a common deployment procedure, showing that the benefit is not solely attributable to the neural decoder.

In the cost-only setting, DFLNOL with neural deployment achieves lower mean, median, P95, and worst-case regret than both Odece variants in all six settings. This demonstrates that training through the continuous representation can provide competitive decisions when uncertainty affects feasibility, without requiring optimal training solutions or repeated combinatorial solver calls. Performance nevertheless deteriorates at the higher polynomial degree, indicating that the framework does not eliminate the difficulty of predicting nonlinear contextual relationships.

Neural deployment achieves lower mean regret than Gurobi deployment in eleven of the twelve settings, although Gurobi retains an advantage on some tail metrics. The exception in mean regret is the 200-node, degree-6 joint-prediction setting: Gurobi reduces mean regret slightly from $27.76\%$ to $27.51\%$ and worst-case regret from $66.99\%$ to $64.70\%$, whereas neural deployment has lower P95 regret ($45.14\%$ versus $45.87\%$). A better solution under predicted parameters need not yield a better executed path under realized costs. These results highlight the importance of evaluating the complete prediction, planning, and execution pipeline.

DFLNOL also has the shortest training time among the evaluated decision-focused methods in every setting, including NOL pretraining. All reported runs finish in under two minutes, whereas Odece requires substantially longer, particularly for larger instances. Neural deployment takes less than 15 milliseconds per instance throughout the table. For the 200-node, degree-6 cost-only setting, it takes approximately 15 milliseconds, compared with approximately 6 seconds for DFLNOL $\rightarrow$ Gurobi. MSE remains cheaper to train but produces substantially higher regret.

The neural-oracle Odece variant provides a further use of the NOL as an approximate optimization oracle. At 200 nodes and degree 6, replacing the online training oracle reduces total training time from approximately 39 hours to 3.4 hours, while mean regret changes from $37.17\%$ to $36.64\%$. The effect on decision quality is not uniform across settings, but training time is reduced in every case. This supports the use of the NOL to reduce repeated solver work within an existing DFL procedure, as well as to support direct differentiation.

\subsection{Sensitivity analysis}

\begin{figure*}[H]
\centering
\includegraphics[width=\textwidth,height=0.83\textheight,keepaspectratio]{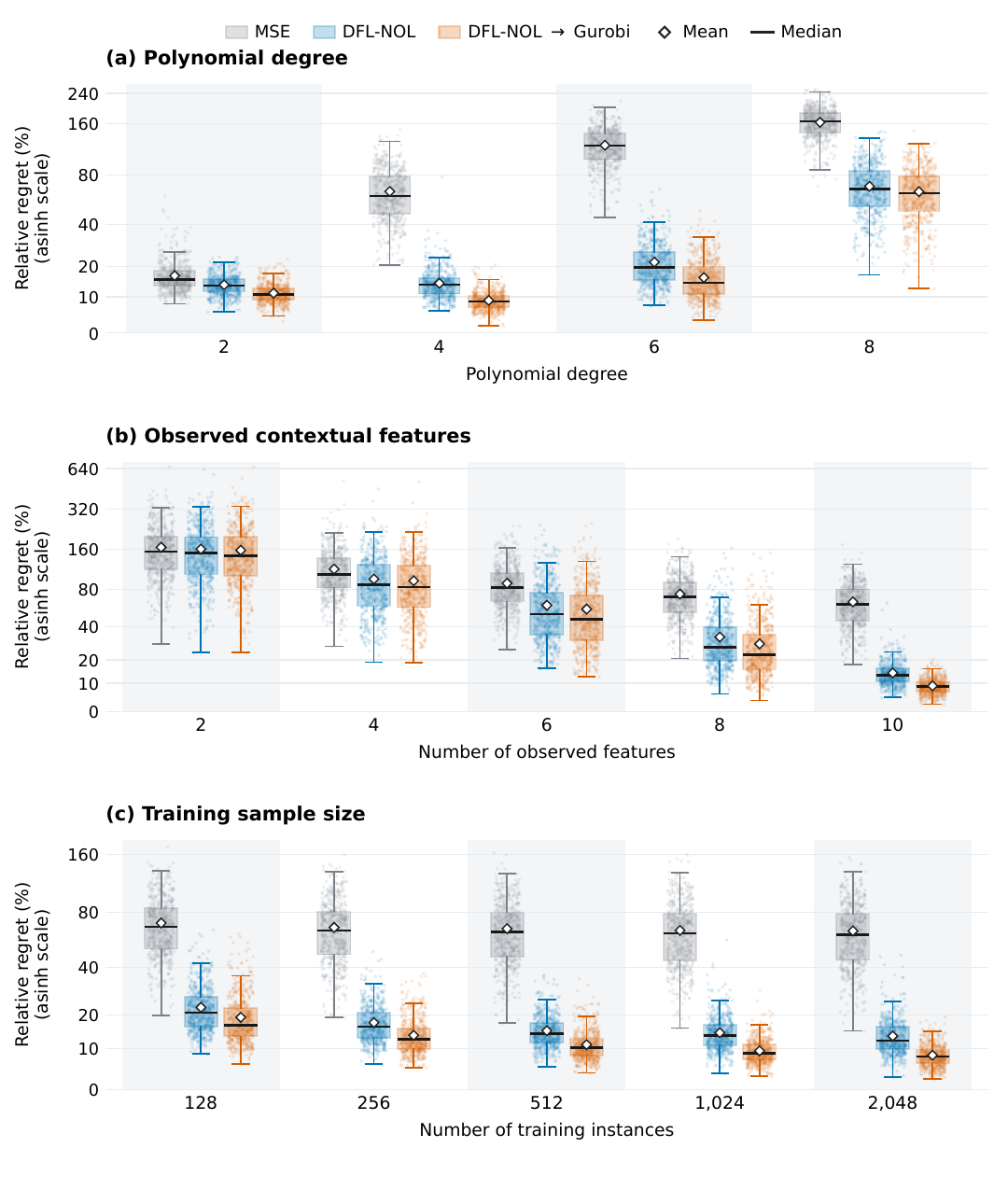}
\caption{Sensitivity analysis on ATSP with 100 nodes. Dots show individual test instances within the displayed ranges. Boxes show interquartile ranges, horizontal lines indicate medians, white diamonds indicate means, and whiskers mark the 5th and 95th percentiles.}
\label{fig:atsp_sensitivity}
\end{figure*}

We examine how decision quality depends on the nonlinearity of the contextual mapping, the availability of contextual information, and the amount of training data. Figure~\ref{fig:atsp_sensitivity} compares MSE, DFLNOL, and DFLNOL $\rightarrow$ Gurobi on 100-node ATSP instances. In panel (a), we vary the polynomial degree over $\{2,4,6,8\}$, using all ten contextual features and 1,000 training instances. In panel (b), we fix the degree at 4 and use 1,000 training instances, but expose only the first $k\in\{2,4,6,8,10\}$ features to the predictor; costs remain generated from all ten features. This separates missing contextual information from changes in the underlying data-generating process. In panel (c), we fix the degree at 4 and retain all ten features while varying the training-set size over $\{128,256,512,1024,2048\}$ to assess data efficiency and diminishing returns. Each setting is evaluated on 500 held-out instances using training seed 0, with the same test instances shared by all three methods within each comparison.

In Figure~\ref{fig:atsp_sensitivity}(a), the regret of MSE increases substantially with the polynomial degree, whereas DFLNOL exhibits a smaller deterioration over degrees 2--6. With a linear predictor, increasing the degree introduces a mismatch between the predictor and the underlying contextual mapping. The results suggest that optimizing predictions for their downstream decision value can mitigate the consequences of this mismatch more effectively than minimizing prediction error alone. Nevertheless, both DFLNOL variants deteriorate noticeably at degree 8: decision-focused training alleviates, but does not eliminate, the limitations of the predictor. Moreover, changing the degree also changes the cost distribution, so this trend should not be attributed solely to increasing nonlinearity.

Figure~\ref{fig:atsp_sensitivity}(b) highlights the role of information availability. When only two contextual features are observed, all three methods exhibit high regret and substantial variability across test instances. As more features become available, regret decreases and the advantage of DFLNOL becomes more pronounced. This suggests that decision-focused training can make effective use of informative context, but cannot fully compensate for missing information. Figure~\ref{fig:atsp_sensitivity}(c) provides a complementary perspective on data availability. DFLNOL already outperforms MSE with 128 training instances, and additional data further improves its performance. However, the improvements become progressively smaller as the training set grows, indicating diminishing marginal returns within the evaluated range rather than a definitive saturation point.

Finally, comparing the two DFLNOL deployment variants helps separate the effect of predictor training from that of deployment. DFLNOL $\to$ Gurobi outperforms MSE despite both using Gurobi at deployment, indicating that the benefit is not solely attributable to the neural decoder. Meanwhile, neural deployment retains much of this improvement without invoking Gurobi. Together, these observations support the usefulness of the NOL both as a differentiable training component and as part of the deployment pipeline.

\section{Conclusion and future development}
\label{sec:conclusion}

This paper presented DFLNOL, a solver-free decision-focused learning framework for edge-selection problems. A self-supervised neural optimization layer provides a continuous decision representation through which predictors learn from objectives and constraint penalties evaluated under realized parameters. The framework accommodates uncertainty in objectives, constraints, or both, without requiring optimal-solution labels or calls to the original NP-hard solver during training and neural deployment. Experiments on ATSP and EFOP demonstrate lower regret than MSE and a favorable trade-off between decision quality and computational cost compared with the evaluated DFL baselines. The layer also reduces training cost when used as an approximate oracle within existing DFL methods.

Future work will focus on improving the agreement between continuous training losses and executed decision quality, particularly when predictor updates shift inputs beyond the optimization layer's training distribution. Extending the framework to additional problem classes and evaluating it on real-world data and larger instances will help establish its broader applicability and robustness.

\bibliographystyle{model1-num-names}

\bibliography{ref.bib}

\begin{thebibliography}{37}
\expandafter\ifx\csname natexlab\endcsname\relax\def\natexlab#1{#1}\fi
\providecommand{\url}[1]{\texttt{#1}}
\providecommand{\href}[2]{#2}
\providecommand{\path}[1]{#1}
\providecommand{\DOIprefix}{doi:}
\providecommand{\ArXivprefix}{arXiv:}
\providecommand{\URLprefix}{URL: }
\providecommand{\Pubmedprefix}{pmid:}
\providecommand{\doi}[1]{\href{http://dx.doi.org/#1}{\path{#1}}}
\providecommand{\Pubmed}[1]{\href{pmid:#1}{\path{#1}}}
\providecommand{\bibinfo}[2]{#2}
\ifx\xfnm\relax \def\xfnm[#1]{\unskip,\space#1}\fi
\bibitem[{Sadana et~al.(2025)Sadana, Chenreddy, Delage, Forel, Frejinger, and Vidal}]{sadana2025survey}
\bibinfo{author}{U.~Sadana}, \bibinfo{author}{A.~Chenreddy}, \bibinfo{author}{E.~Delage}, \bibinfo{author}{A.~Forel}, \bibinfo{author}{E.~Frejinger}, \bibinfo{author}{T.~Vidal},
\newblock \bibinfo{title}{A survey of contextual optimization methods for decision-making under uncertainty},
\newblock \bibinfo{journal}{European Journal of Operational Research} \bibinfo{volume}{320} (\bibinfo{year}{2025}) \bibinfo{pages}{271--289}.
\bibitem[{Mandi et~al.(2024)Mandi, Kotary, Berden, Mulamba, Bucarey, Guns, and Fioretto}]{mandi2024decision}
\bibinfo{author}{J.~Mandi}, \bibinfo{author}{J.~Kotary}, \bibinfo{author}{S.~Berden}, \bibinfo{author}{M.~Mulamba}, \bibinfo{author}{V.~Bucarey}, \bibinfo{author}{T.~Guns}, \bibinfo{author}{F.~Fioretto},
\newblock \bibinfo{title}{Decision-focused learning: Foundations, state of the art, benchmark and future opportunities},
\newblock \bibinfo{journal}{Journal of Artificial Intelligence Research} \bibinfo{volume}{80} (\bibinfo{year}{2024}) \bibinfo{pages}{1623--1701}.
\bibitem[{Wilder et~al.(2019)Wilder, Dilkina, and Tambe}]{wilder2019melding}
\bibinfo{author}{B.~Wilder}, \bibinfo{author}{B.~Dilkina}, \bibinfo{author}{M.~Tambe},
\newblock \bibinfo{title}{Melding the data-decisions pipeline: Decision-focused learning for combinatorial optimization},
\newblock in: \bibinfo{booktitle}{Proceedings of the AAAI Conference on Artificial Intelligence}, volume~\bibinfo{volume}{33}, \bibinfo{year}{2019}, pp. \bibinfo{pages}{1658--1665}. \DOIprefix\doi{10.1609/aaai.v33i01.33011658}.
\bibitem[{Mandi et~al.(2020)Mandi, Demirovi{\'c}, Stuckey, and Guns}]{mandi2020smart}
\bibinfo{author}{J.~Mandi}, \bibinfo{author}{E.~Demirovi{\'c}}, \bibinfo{author}{P.~J. Stuckey}, \bibinfo{author}{T.~Guns},
\newblock \bibinfo{title}{Smart predict-and-optimize for hard combinatorial optimization problems},
\newblock in: \bibinfo{booktitle}{Proceedings of the AAAI Conference on Artificial Intelligence}, volume~\bibinfo{volume}{34}, \bibinfo{year}{2020}, pp. \bibinfo{pages}{1603--1610}. \DOIprefix\doi{10.1609/aaai.v34i02.5521}.
\bibitem[{Vlastelica et~al.(2020)Vlastelica, Paulus, Musil, Martius, and Rol{\'i}nek}]{vlastelica2020differentiation}
\bibinfo{author}{M.~Vlastelica}, \bibinfo{author}{A.~Paulus}, \bibinfo{author}{V.~Musil}, \bibinfo{author}{G.~Martius}, \bibinfo{author}{M.~Rol{\'i}nek},
\newblock \bibinfo{title}{Differentiation of blackbox combinatorial solvers},
\newblock in: \bibinfo{booktitle}{International Conference on Learning Representations}, \bibinfo{year}{2020}. \DOIprefix\doi{10.48550/arXiv.1912.02175}.
\bibitem[{Tang and Khalil(2024)}]{tang2024cave}
\bibinfo{author}{B.~Tang}, \bibinfo{author}{E.~B. Khalil},
\newblock \bibinfo{title}{{CaVE}: A cone-aligned approach for fast predict-then-optimize with binary linear programs},
\newblock in: \bibinfo{booktitle}{Integration of Constraint Programming, Artificial Intelligence, and Operations Research (CPAIOR 2024)}, volume \bibinfo{volume}{14743} of \textit{\bibinfo{series}{Lecture Notes in Computer Science}}, \bibinfo{year}{2024}, pp. \bibinfo{pages}{193--210}. \DOIprefix\doi{10.1007/978-3-031-60599-4_12}.
\bibitem[{Berden et~al.(2025)Berden, Mahmuto{\u{g}}ullar{\i}, Tsouros, and Guns}]{berden2025solverfree}
\bibinfo{author}{S.~Berden}, \bibinfo{author}{A.~{\.I}. Mahmuto{\u{g}}ullar{\i}}, \bibinfo{author}{D.~Tsouros}, \bibinfo{author}{T.~Guns},
\newblock \bibinfo{title}{Solver-free decision-focused learning for linear optimization problems},
\newblock in: \bibinfo{booktitle}{Advances in Neural Information Processing Systems}, volume~\bibinfo{volume}{38}, \bibinfo{year}{2025}, pp. \bibinfo{pages}{158127--158145}. \DOIprefix\doi{10.52202/085713-5284}.
\bibitem[{Kotary et~al.(2024)Kotary, Di~Vito, Christopher, Van~Hentenryck, and Fioretto}]{kotary2024learning}
\bibinfo{author}{J.~Kotary}, \bibinfo{author}{V.~Di~Vito}, \bibinfo{author}{J.~Christopher}, \bibinfo{author}{P.~Van~Hentenryck}, \bibinfo{author}{F.~Fioretto},
\newblock \bibinfo{title}{Learning joint models of prediction and optimization},
\newblock in: \bibinfo{booktitle}{ECAI 2024 -- 27th European Conference on Artificial Intelligence}, volume \bibinfo{volume}{392} of \textit{\bibinfo{series}{Frontiers in Artificial Intelligence and Applications}}, \bibinfo{publisher}{IOS Press}, \bibinfo{year}{2024}, pp. \bibinfo{pages}{2476--2483}. \DOIprefix\doi{10.3233/FAIA240775}.
\bibitem[{Elmachtoub and Grigas(2022)}]{elmachtoub2022smart}
\bibinfo{author}{A.~N. Elmachtoub}, \bibinfo{author}{P.~Grigas},
\newblock \bibinfo{title}{Smart ``predict, then optimize''},
\newblock \bibinfo{journal}{Management Science} \bibinfo{volume}{68} (\bibinfo{year}{2022}) \bibinfo{pages}{9--26}.
\bibitem[{Hu et~al.(2023)Hu, Lee, and Lee}]{hu2023twostage}
\bibinfo{author}{X.~Hu}, \bibinfo{author}{J.~C.~H. Lee}, \bibinfo{author}{J.~H.~M. Lee},
\newblock \bibinfo{title}{Two-stage predict+{O}ptimize for {MILP}s with unknown parameters in constraints},
\newblock in: \bibinfo{booktitle}{Advances in Neural Information Processing Systems}, volume~\bibinfo{volume}{36}, \bibinfo{year}{2023}, pp. \bibinfo{pages}{14247--14272}. \URLprefix \url{https://papers.neurips.cc/paper_files/paper/2023/file/2e14be0332c04c76742710e417cedb2a-Paper-Conference.pdf}. \DOIprefix\doi{10.52202/075280-0627}.
\bibitem[{Mandi et~al.(2025)Mandi, Defresne, Berden, and Guns}]{mandi2025odece}
\bibinfo{author}{J.~Mandi}, \bibinfo{author}{M.~Defresne}, \bibinfo{author}{S.~Berden}, \bibinfo{author}{T.~Guns},
\newblock \bibinfo{title}{Feasibility-aware decision-focused learning for predicting parameters in the constraints},
\newblock in: \bibinfo{booktitle}{Advances in Neural Information Processing Systems}, volume~\bibinfo{volume}{38}, \bibinfo{year}{2025}, pp. \bibinfo{pages}{163285--163309}. \DOIprefix\doi{10.52202/085713-5451}.
\bibitem[{Cristian et~al.(2023)Cristian, Harsha, Perakis, Quanz, and Spantidakis}]{cristian2023endtoend}
\bibinfo{author}{R.~Cristian}, \bibinfo{author}{P.~Harsha}, \bibinfo{author}{G.~Perakis}, \bibinfo{author}{B.~L. Quanz}, \bibinfo{author}{I.~Spantidakis},
\newblock \bibinfo{title}{End-to-end learning for optimization via constraint-enforcing approximators},
\newblock in: \bibinfo{booktitle}{Proceedings of the AAAI Conference on Artificial Intelligence}, volume~\bibinfo{volume}{37}, \bibinfo{year}{2023}, pp. \bibinfo{pages}{7253--7260}. \DOIprefix\doi{10.1609/aaai.v37i6.25884}.
\bibitem[{Cristian et~al.(2025)Cristian, Harsha, Perakis, and Quanz}]{cristian2025efficient}
\bibinfo{author}{R.~Cristian}, \bibinfo{author}{P.~Harsha}, \bibinfo{author}{G.~Perakis}, \bibinfo{author}{B.~Quanz}, \bibinfo{title}{Efficient end-to-end learning for decision-making: A meta-optimization approach}, \bibinfo{year}{2025}. \DOIprefix\doi{10.48550/arXiv.2505.11360}. \href{http://arxiv.org/abs/2505.11360}{\tt arXiv:2505.11360}.
\bibitem[{Liu et~al.(2022)Liu, Wang, Cao, Liu, and Shen}]{liu2022gon}
\bibinfo{author}{C.~Liu}, \bibinfo{author}{J.~Wang}, \bibinfo{author}{Y.~Cao}, \bibinfo{author}{M.~Liu}, \bibinfo{author}{W.~Shen},
\newblock \bibinfo{title}{{GON}: End-to-end optimization framework for constraint graph optimization problems},
\newblock \bibinfo{journal}{Knowledge-Based Systems} \bibinfo{volume}{254} (\bibinfo{year}{2022}) \bibinfo{pages}{109697}.
\bibitem[{Zheng et~al.(2026)Zheng, de~Nijs, and Lam}]{zheng2026constraint}
\bibinfo{author}{X.~Zheng}, \bibinfo{author}{F.~de~Nijs}, \bibinfo{author}{E.~Lam},
\newblock \bibinfo{title}{Constraint-aware self-supervised learning for edge selection},
\newblock in: \bibinfo{booktitle}{32nd International Conference on Principles and Practice of Constraint Programming (CP 2026)}, volume \bibinfo{volume}{379} of \textit{\bibinfo{series}{Leibniz International Proceedings in Informatics (LIPIcs)}}, \bibinfo{year}{2026}, pp. \bibinfo{pages}{61:1--61:18}. \DOIprefix\doi{10.4230/LIPIcs.CP.2026.61}.
\bibitem[{Wang et~al.(2022)Wang, Wu, Yang, Hao, and Li}]{wang2022unsupervised}
\bibinfo{author}{H.~P. Wang}, \bibinfo{author}{N.~Wu}, \bibinfo{author}{H.~Yang}, \bibinfo{author}{C.~Hao}, \bibinfo{author}{P.~Li},
\newblock \bibinfo{title}{Unsupervised learning for combinatorial optimization with principled objective relaxation},
\newblock in: \bibinfo{booktitle}{Advances in Neural Information Processing Systems}, volume~\bibinfo{volume}{35}, \bibinfo{year}{2022}, pp. \bibinfo{pages}{31444--31458}. \DOIprefix\doi{10.52202/068431-2280}.
\bibitem[{Amos and Kolter(2017)}]{amos2017optnet}
\bibinfo{author}{B.~Amos}, \bibinfo{author}{J.~Z. Kolter},
\newblock \bibinfo{title}{{OptNet}: Differentiable optimization as a layer in neural networks},
\newblock in: \bibinfo{booktitle}{Proceedings of the 34th International Conference on Machine Learning}, volume~\bibinfo{volume}{70} of \textit{\bibinfo{series}{Proceedings of Machine Learning Research}}, \bibinfo{year}{2017}, pp. \bibinfo{pages}{136--145}. \DOIprefix\doi{10.48550/arXiv.1703.00443}.
\bibitem[{Agrawal et~al.(2019)Agrawal, Amos, Barratt, Boyd, Diamond, and Kolter}]{agrawal2019differentiable}
\bibinfo{author}{A.~Agrawal}, \bibinfo{author}{B.~Amos}, \bibinfo{author}{S.~Barratt}, \bibinfo{author}{S.~Boyd}, \bibinfo{author}{S.~Diamond}, \bibinfo{author}{J.~Z. Kolter},
\newblock \bibinfo{title}{Differentiable convex optimization layers},
\newblock in: \bibinfo{booktitle}{Advances in Neural Information Processing Systems}, volume~\bibinfo{volume}{32}, \bibinfo{year}{2019}. \DOIprefix\doi{10.48550/arXiv.1910.12430}.
\bibitem[{Ferber et~al.(2020)Ferber, Wilder, Dilkina, and Tambe}]{ferber2020mipaal}
\bibinfo{author}{A.~Ferber}, \bibinfo{author}{B.~Wilder}, \bibinfo{author}{B.~Dilkina}, \bibinfo{author}{M.~Tambe},
\newblock \bibinfo{title}{{MIPaaL}: Mixed integer program as a layer},
\newblock in: \bibinfo{booktitle}{Proceedings of the AAAI Conference on Artificial Intelligence}, volume~\bibinfo{volume}{34}, \bibinfo{year}{2020}, pp. \bibinfo{pages}{1504--1511}. \DOIprefix\doi{10.1609/aaai.v34i02.5509}.
\bibitem[{McKenzie et~al.(2024)McKenzie, Wu~Fung, and Heaton}]{mckenzie2024differentiating}
\bibinfo{author}{D.~McKenzie}, \bibinfo{author}{S.~Wu~Fung}, \bibinfo{author}{H.~Heaton},
\newblock \bibinfo{title}{Differentiating through integer linear programs with quadratic regularization and {Davis--Yin} splitting},
\newblock \bibinfo{journal}{Transactions on Machine Learning Research}  (\bibinfo{year}{2024}).
\bibitem[{Mandi et~al.(2025)Mandi, Mahmuto{\u{g}}ullar{\i}, Berden, and Guns}]{mandi2025surrogate}
\bibinfo{author}{J.~Mandi}, \bibinfo{author}{A.~{\.I}. Mahmuto{\u{g}}ullar{\i}}, \bibinfo{author}{S.~Berden}, \bibinfo{author}{T.~Guns},
\newblock \bibinfo{title}{Minimizing surrogate losses for decision-focused learning using differentiable optimization},
\newblock in: \bibinfo{booktitle}{ECAI 2025}, \bibinfo{year}{2025}. \DOIprefix\doi{10.3233/FAIA251273}.
\bibitem[{Berthet et~al.(2020)Berthet, Blondel, Teboul, Cuturi, Vert, and Bach}]{berthet2020learning}
\bibinfo{author}{Q.~Berthet}, \bibinfo{author}{M.~Blondel}, \bibinfo{author}{O.~Teboul}, \bibinfo{author}{M.~Cuturi}, \bibinfo{author}{J.-P. Vert}, \bibinfo{author}{F.~Bach},
\newblock \bibinfo{title}{Learning with differentiable perturbed optimizers},
\newblock in: \bibinfo{booktitle}{Advances in Neural Information Processing Systems}, volume~\bibinfo{volume}{33}, \bibinfo{year}{2020}, pp. \bibinfo{pages}{9508--9519}. \DOIprefix\doi{10.48550/arXiv.2002.08676}.
\bibitem[{Gupta and Huang(2024)}]{gupta2024directional}
\bibinfo{author}{V.~Gupta}, \bibinfo{author}{M.~Huang},
\newblock \bibinfo{title}{Decision-focused learning with directional gradients},
\newblock in: \bibinfo{booktitle}{Advances in Neural Information Processing Systems}, volume~\bibinfo{volume}{37}, \bibinfo{year}{2024}, pp. \bibinfo{pages}{79194--79220}. \URLprefix \url{https://proceedings.neurips.cc/paper_files/paper/2024/hash/907a9fb75a408f6c3a2ae1bf84c39e44-Abstract-Conference.html}. \DOIprefix\doi{10.52202/079017-2514}.
\bibitem[{Mulamba et~al.(2021)Mulamba, Mandi, Diligenti, Lombardi, Bucarey, and Guns}]{mulamba2021contrastive}
\bibinfo{author}{M.~Mulamba}, \bibinfo{author}{J.~Mandi}, \bibinfo{author}{M.~Diligenti}, \bibinfo{author}{M.~Lombardi}, \bibinfo{author}{V.~Bucarey}, \bibinfo{author}{T.~Guns},
\newblock \bibinfo{title}{Contrastive losses and solution caching for predict-and-optimize},
\newblock in: \bibinfo{booktitle}{Proceedings of the Thirtieth International Joint Conference on Artificial Intelligence}, \bibinfo{year}{2021}, pp. \bibinfo{pages}{2833--2840}. \DOIprefix\doi{10.24963/ijcai.2021/390}.
\bibitem[{Mandi et~al.(2022)Mandi, Bucarey, Tchomba, and Guns}]{mandi2022decision}
\bibinfo{author}{J.~Mandi}, \bibinfo{author}{V.~Bucarey}, \bibinfo{author}{M.~M.~K. Tchomba}, \bibinfo{author}{T.~Guns},
\newblock \bibinfo{title}{Decision-focused learning: Through the lens of learning to rank},
\newblock in: \bibinfo{booktitle}{Proceedings of the 39th International Conference on Machine Learning}, volume \bibinfo{volume}{162} of \textit{\bibinfo{series}{Proceedings of Machine Learning Research}}, \bibinfo{year}{2022}, pp. \bibinfo{pages}{14935--14947}. \URLprefix \url{https://proceedings.mlr.press/v162/mandi22a.html}. \DOIprefix\doi{10.48550/arXiv.2112.03609}.
\bibitem[{Shah et~al.(2022)Shah, Wang, Wilder, Perrault, and Tambe}]{shah2022decision}
\bibinfo{author}{S.~Shah}, \bibinfo{author}{K.~Wang}, \bibinfo{author}{B.~Wilder}, \bibinfo{author}{A.~Perrault}, \bibinfo{author}{M.~Tambe},
\newblock \bibinfo{title}{Decision-focused learning without decision-making: Learning locally optimized decision losses},
\newblock in: \bibinfo{booktitle}{Advances in Neural Information Processing Systems}, volume~\bibinfo{volume}{35}, \bibinfo{year}{2022}, pp. \bibinfo{pages}{1320--1332}. \DOIprefix\doi{10.52202/068431-0097}.
\bibitem[{Shah et~al.(2024)Shah, Wilder, Perrault, and Tambe}]{shah2024leaving}
\bibinfo{author}{S.~Shah}, \bibinfo{author}{B.~Wilder}, \bibinfo{author}{A.~Perrault}, \bibinfo{author}{M.~Tambe},
\newblock \bibinfo{title}{Leaving the nest: Going beyond local loss functions for predict-then-optimize},
\newblock in: \bibinfo{booktitle}{Proceedings of the AAAI Conference on Artificial Intelligence}, volume~\bibinfo{volume}{38}, \bibinfo{year}{2024}, pp. \bibinfo{pages}{14902--14909}. \DOIprefix\doi{10.1609/aaai.v38i13.29410}.
\bibitem[{Zharmagambetov et~al.(2023)Zharmagambetov, Amos, Ferber, Huang, Dilkina, and Tian}]{zharmagambetov2023landscape}
\bibinfo{author}{A.~Zharmagambetov}, \bibinfo{author}{B.~Amos}, \bibinfo{author}{A.~Ferber}, \bibinfo{author}{T.~Huang}, \bibinfo{author}{B.~Dilkina}, \bibinfo{author}{Y.~Tian},
\newblock \bibinfo{title}{Landscape surrogate: Learning decision losses for mathematical optimization under partial information},
\newblock in: \bibinfo{booktitle}{Advances in Neural Information Processing Systems}, volume~\bibinfo{volume}{36}, \bibinfo{year}{2023}, pp. \bibinfo{pages}{27332--27350}. \DOIprefix\doi{10.52202/075280-1191}.
\bibitem[{Hu et~al.(2023)Hu, Lee, and Lee}]{hu2023packing}
\bibinfo{author}{X.~Hu}, \bibinfo{author}{J.~C.~H. Lee}, \bibinfo{author}{J.~H.~M. Lee},
\newblock \bibinfo{title}{Predict+optimize for packing and covering {LPs} with unknown parameters in constraints},
\newblock in: \bibinfo{booktitle}{Proceedings of the AAAI Conference on Artificial Intelligence}, volume~\bibinfo{volume}{37}, \bibinfo{year}{2023}, pp. \bibinfo{pages}{3987--3995}. \DOIprefix\doi{10.1609/aaai.v37i4.25513}.
\bibitem[{Donti et~al.(2021)Donti, Rolnick, and Kolter}]{donti2021dc3}
\bibinfo{author}{P.~L. Donti}, \bibinfo{author}{D.~Rolnick}, \bibinfo{author}{J.~Z. Kolter},
\newblock \bibinfo{title}{{DC3}: A learning method for optimization with hard constraints},
\newblock in: \bibinfo{booktitle}{International Conference on Learning Representations (ICLR)}, \bibinfo{year}{2021}. \DOIprefix\doi{10.48550/arXiv.2104.12225}. \href{http://arxiv.org/abs/2104.12225}{\tt arXiv:2104.12225}.
\bibitem[{Park and Van~Hentenryck(2023)}]{park2023selfsupervised}
\bibinfo{author}{S.~Park}, \bibinfo{author}{P.~Van~Hentenryck},
\newblock \bibinfo{title}{Self-supervised primal-dual learning for constrained optimization},
\newblock in: \bibinfo{booktitle}{Proceedings of the AAAI Conference on Artificial Intelligence}, volume~\bibinfo{volume}{37}, \bibinfo{year}{2023}, pp. \bibinfo{pages}{4052--4060}. \DOIprefix\doi{10.1609/aaai.v37i4.25520}.
\bibitem[{Sun and Yang(2023)}]{sun2023difusco}
\bibinfo{author}{Z.~Sun}, \bibinfo{author}{Y.~Yang},
\newblock \bibinfo{title}{{DIFUSCO}: Graph-based diffusion solvers for combinatorial optimization},
\newblock in: \bibinfo{booktitle}{Advances in Neural Information Processing Systems}, volume~\bibinfo{volume}{36}, \bibinfo{year}{2023}, pp. \bibinfo{pages}{3706--3731}. \DOIprefix\doi{10.52202/075280-0164}.
\bibitem[{Kool et~al.(2019)Kool, van Hoof, and Welling}]{kool2019attention}
\bibinfo{author}{W.~Kool}, \bibinfo{author}{H.~van Hoof}, \bibinfo{author}{M.~Welling},
\newblock \bibinfo{title}{Attention, learn to solve routing problems!},
\newblock in: \bibinfo{booktitle}{International Conference on Learning Representations}, \bibinfo{year}{2019}. \DOIprefix\doi{10.48550/arXiv.1803.08475}.
\bibitem[{Qiu et~al.(2022)Qiu, Sun, and Yang}]{qiu2022dimes}
\bibinfo{author}{R.~Qiu}, \bibinfo{author}{Z.~Sun}, \bibinfo{author}{Y.~Yang},
\newblock \bibinfo{title}{{DIMES}: A differentiable meta solver for combinatorial optimization problems},
\newblock in: \bibinfo{booktitle}{Advances in Neural Information Processing Systems}, volume~\bibinfo{volume}{35}, \bibinfo{year}{2022}, pp. \bibinfo{pages}{25531--25546}. \DOIprefix\doi{10.52202/068431-1851}.
\bibitem[{Tang and Khalil(2024)}]{tang2024pyepo}
\bibinfo{author}{B.~Tang}, \bibinfo{author}{E.~B. Khalil},
\newblock \bibinfo{title}{{PyEPO}: A {PyTorch}-based end-to-end predict-then-optimize library for linear and integer programming},
\newblock \bibinfo{journal}{Mathematical Programming Computation} \bibinfo{volume}{16} (\bibinfo{year}{2024}) \bibinfo{pages}{297--335}.
\bibitem[{{Gurobi Optimization, LLC}(2024)}]{gurobi}
\bibinfo{author}{{Gurobi Optimization, LLC}}, \bibinfo{title}{Gurobi optimizer reference manual}, \bibinfo{year}{2024}. \URLprefix \url{https://www.gurobi.com}.
\bibitem[{Roberti and Toth(2012)}]{roberti2012models}
\bibinfo{author}{R.~Roberti}, \bibinfo{author}{P.~Toth},
\newblock \bibinfo{title}{Models and algorithms for the {Asymmetric Traveling Salesman Problem}: an experimental comparison},
\newblock \bibinfo{journal}{EURO Journal on Transportation and Logistics} \bibinfo{volume}{1} (\bibinfo{year}{2012}) \bibinfo{pages}{113--133}.

\end{thebibliography}

\end{document}